\documentclass[12pt,a4paper]{elsarticle}
\usepackage[utf8]{inputenc}
\usepackage[T1]{fontenc}
\usepackage{amsmath,amsthm}
\usepackage{amsfonts}
\usepackage{amssymb}
\usepackage{graphicx,subfigure,xcolor}
\usepackage[ruled,vlined]{algorithm2e}
\usepackage{tikz}
\usetikzlibrary{matrix,shapes,arrows,positioning}
\usepackage[margin=0.99in]{geometry}    % Margin: 1-inch all round

\makeatletter
\def\ps@pprintTitle{%
  \let\@oddhead\@empty
  \let\@evenhead\@empty
  \let\@oddfoot\@empty
  \let\@evenfoot\@oddfoot
}
\makeatother

\begin{document}
\begin{frontmatter}  
\title{DeepParticle: learning invariant measure by a deep neural network minimizing Wasserstein distance on data generated from an interacting particle method}
       \author[uoc]{Zhongjian Wang}
        \ead{zhongjian@statistics.uchicago.edu}
		\author[uci]{Jack Xin}
		\ead{jxin@math.uci.edu}
		\author[hku]{Zhiwen Zhang\corref{cor1}}
		\ead{zhangzw@hku.hk}
		
		\address[uoc]{Department of Statistics and CCAM, The University of Chicago, Chicago, IL 60637, USA.}
		\address[uci]{Department of Mathematics, University of California at Irvine, Irvine, CA 92697, USA.}
		\address[hku]{Department of Mathematics, The University of Hong Kong, Pokfulam Road, Hong Kong SAR, China.}		
		\cortext[cor1]{Corresponding author}
\begin{abstract}
	We introduce DeepParticle, a method to learn and generate invariant measures of stochastic dynamical systems with physical parameters based on data computed from an interacting particle method (IPM). We utilize the expressiveness of deep neural networks (DNNs) to represent the transform of samples from a given input (source) distribution to an arbitrary target distribution, neither assuming distribution functions in closed form nor the sample transforms to be invertible, nor the sample state space to be finite. In the training stage, we update the weights of the network by minimizing a discrete Wasserstein distance between the input and target samples. To reduce the computational cost, we propose an iterative divide-and-conquer (a mini-batch interior point) algorithm, to find the optimal transition matrix in the Wasserstein distance. We present numerical results to demonstrate the performance of our method for accelerating the computation of invariant measures of stochastic dynamical systems arising in computing reaction-diffusion front speeds in 3D chaotic flows by using the  IPM.  The physical parameter is a large P\'eclet number reflecting the advection-dominated regime of our interest.%\\
	
	\noindent \textit{\textbf{AMS subject classification:}} 35K57, 37M25, 49Q22, 65C35, 68T07. \\
	
	\noindent \textit{\textbf{Keywords:}}  physically parameterized invariant measures, interacting particle method, deep learning, Wasserstein distance, front speeds in chaotic flows.
\end{abstract}
\end{frontmatter}
\setcounter{page}{1}
\section{Introduction}

\noindent	 
In recent years, deep neural networks (DNNs) have achieved unprecedented levels of success in a broad range of areas such as computer vision, speech recognition, natural language processing, and health sciences, producing results comparable or superior to human experts \cite{lecun2015deep,goodfellow2016deep}. The impacts have reached physical sciences where traditional first-principle based modeling and computational methodologies have been the norm. Thanks in part to the user-friendly open-source computing platforms from industry (e.g. TensorFlow and PyTorch), there have been vibrant activities in applying deep learning tools for scientific computing, such as approximating multivariate functions, solving ordinary/partial differential equations (ODEs/PDEs) and inverse problems using DNNs; see e.g. \cite{Kutz_17,Kutz_19,weinan2017deep,weinan2018deep,JinchaoXu:2018,QiangDU:2018,khoo2017solving,Brenner_18,ZabarasZhu:2018,Karniadakis2018learning,sirignano2018dgm,zhu2019physics,raissi2019physics,karumuri2019simulator,Dongb_18,Dongb_19,Xiu_19,Xiu_20,BaoZhou_20a,BaoZhou_20b,wang2020mesh,Shen_21a,Shen_21b,yoo2020deep} and references therein.  
\medskip

There are many classical works on the approximation power of neural networks; see e.g. \cite{cybenko1989approximation,hornik1989multilayer,ellacott1994aspects,pinkus1999approximation}.  For recent works on the expressive (approximation) power of DNNs; see e.g. \cite{cohen2016expressive,schwab2017deep,yarotsky2017error,QiangDU:2018,Shen_21a,Shen_21b}. In \cite{JinchaoXu:2018}, the authors 
%investigate the relationship between DNNs with rectified linear unit (ReLU) function as the activation function and continuous piece-wise linear basis functions in finite element method (FEM). They 
showed that DNNs with rectified linear unit (ReLU) activation function and enough width/depth contain the continuous piece-wise linear  finite element space. %Thus, one can represent a solution of PDE using the ReLU-DNN. 
\medskip

Solving ODEs or PDEs by a neural network (NN) approximation is known in the literature dating back at least to the 1990's; see e.g. \cite{LeeH1990, Fernandez1994, Lagaris1998}. The main idea in these works is to train NNs to approximate the solution by minimizing the residual of the ODEs or PDEs, along with the associated initial and boundary conditions. These early works estimate neural network solutions on a fixed mesh, however.
Recently 
%in \cite{weinan2018deep}, DNNs are proposed for Poisson and eigenvalue problems with variational principle (deep Ritz), and in \cite{han2018solving} for a class of high-dimensional parabolic PDEs with stochastic representations. 
DNN methods are developed for Poisson and eigenvalue problems with a variational principle characterization (deep Ritz,  \cite{weinan2018deep}), for a class of high-dimensional parabolic PDEs with stochastic representations \cite{han2018solving}, and for advancing finite element methods \cite{Xu_mgnet,Cai_20,Cai_21}.
The physics-informed neural network (PINN) method \cite{raissi2019physics} and a deep Galerkin method (DGM) \cite{sirignano2018dgm} compute PDE solutions based on their physical properties. For parametric PDEs, 
a deep operator network (DeepONet) learns operators accurately and efficiently from a relatively small dataset based on the universal approximation theorem of operators \cite{lu2019deeponet};
a Fourier neural operator method \cite{li2020fourier}  directly learns the mapping from functional parametric dependence to the solutions (of a family of PDEs). 
%in contrast to methods that solve one instance of an equation. Meanwhile, a deep operator network (DeepONet) was proposed to learn operators accurately and efficiently from a relatively small dataset based on the universal approximation theorem of operators \cite{lu2019deeponet}. 
Deep basis learning is studied in \cite{JackXinLyu:2020} to improve proper orthogonal decomposition for residual diffusivity in chaotic flows \cite{JackXinLyu:2017}, among other works on reduced order modeling \cite{Kutz_17,Hest_20,Hest_21}. 
In \cite{BaoZhou_20a,BaoZhou_20b}, weak 
 adversarial network methods are studied for weak solutions and inverse problems, see also related studies on PDE recovery from data via DNN \cite{Dongb_18,Dongb_19,Xiu_19, Xiu_20} among others. 
In the context of surrogate modeling and uncertainty quantification (UQ), DNN methods include Bayesian deep convolutional encoder-decoder networks \cite{ZabarasZhu:2018}, deep multi-scale model learning \cite{Yalchin2018deep}, physics-constrained deep learning method \cite{zhu2019physics}, see also \cite{khoo2017solving,schwab2017deep,karumuri2019simulator,Karn2021} and references therein. For DNN applications in mean field games (high dimensional optimal control problems) and various connections with numerical PDEs, see \cite{Carmona_MFG1,Carmona_MFG2,Osher_20,Osher_21,Osher_sMFG_21} and references therein. In view of the above literature, the DNN interactions with numerical PDE methods mostly occur in the Eulerian setting with PDE solutions defined in either the strong or weak (variational) sense. 
	%
	%In \cite{khoo2017solving}, a neural network was proposed to learn the physical quantity of interest as a function of random input coefficients; the accuracy and efficiency of the approach for solving parametric PDE problems was shown. In \cite{ZabarasZhu:2018}, the authors propose a Bayesian approach to develop deep convolutional encoder-decoder networks, which give surrogate models for uncertainty quantification and propagation in problems governed by stochastic PDEs. In \cite{schwab2017deep}, the authors estimate the expressive power of a class of deep Neural Networks on a class of countably-parametric maps. Those maps arise as response surfaces of parametric PDEs with distributed uncertain inputs. In \cite{Yalchin2018deep}, the authors design multi-layer neural network architectures for multiscale simulations of flows that takes into account the observed data and physical modeling concepts.  In \cite{Karniadakis2018learning}, the authors apply the multistep neural networks to identify nonlinear dynamical systems from observed data. They test the effectiveness of the proposed approach for several benchmark problems. 
	
	\medskip
	
Our goal here is to study {\it deep learning in a Lagrangian framework of multi-scale PDE problems}, coming naturally from our recent work (\cite{IPM_2021}, reviewed in section 4.2 later) on a convergent interacting particle method (IPM) for computing large-scale reaction-diffusion front speeds through chaotic flows. The method is based on a genetic algorithm that evolves a large ensemble of uniformly distributed particles at the initial time to another ensemble of particles obeying an invariant measure at a large time. The front speed is readily computed from the invariant measure. Though the method is mesh-free, the computational costs remain high as advection becomes dominant at large P\'eclet numbers.  Clearly, it is desirable to initialize particle distribution with some resemblance of the target invariant measure instead of starting from the uniform distribution. Hence learning from samples of invariant measure at a smaller P\'eclet number with more affordable computation to predict an invariant measure at a larger P\'eclet number becomes a natural problem to study. 
	\medskip
	
Specifically, we shall develop a DNN $f(\cdot; \theta)$ to map a uniform distribution $\mu$ (source) to an invariant measure $\nu = \nu(\kappa)$ (target),  where $\kappa$ is the reciprocal of P\'eclet number, and $\theta$ consists of network weights and $\kappa$.  The network is deep from input to output with $12$ Sigmoid layers while the first three layers are coupled to a shallow companion network to account for the effects of parameter $\kappa$ on the network weights. In addition, we include both local and nonlocal skip connections along the deep direction to assist information flow. The network is trained by minimizing the 2-Wasserstein distance (2-WD) between two measures $\mu$ and $\nu$ \cite{villani2021topics}. We consider a discrete version of 2-WD for finitely many samples of $\mu$ and $\nu$, which involves a linear program (LP) optimizing over doubly stochastic matrices \cite{sinkhorn1964relationship}. Directly solving the LP by the interior point method \cite{wright1997primal} is too costly. We devise a mini-batch interior point method by sampling smaller sub-matrices while preserving row and column sums.  This turns out to be very efficient and integrated well with the stochastic gradient descent (SGD) method for the entire network training. 
\medskip
	  
We shall conduct three numerical experiments to verify the performance of the proposed DeepParticle method. The first example is a synthetic data set on $\mathbb{R}^1$, where $\mu$ is a uniform distribution and $\nu$ is a normal distribution with zero mean and variance $\sigma^{2}_{1}$. In the second example, we compute the Kolmogorov-Petrovsky-Piskunov (KPP) front speed in a 2D steady cellular flow and learn the invariant measures corresponding to different $\kappa$'s. In this experiment, $\mu$ is a uniform distribution on $[0, 2\pi]^2$, and $\nu$ an empirical invariant measure obtained from IPM simulation of reaction-diffusion particles in advecting flows with P\'eclet number $O(\kappa^{-1})$. Finally, we compute the KPP front speed in a 3D time-dependent Kolmogorov flow with chaotic streamlines and study the IPM evolution of the measure in three-dimensional space. In this experiment, $\mu$ is  a uniform distribution on $[0, 2\pi]^3$, and $\nu$ is an empirical invariant measure obtained from IPM simulation of reaction-diffusion particles of the KPP equation in time-dependent Kolmogorov flow. Numerical results show that  the proposed DeepParticle method efficiently learns the $\kappa$ dependent target distributions and predicts the physically meaningful sharpening effect as $\kappa$ becomes small. 
\medskip

We remark that though there are other techniques for mapping distributions such as Normalizing Flows (NF) \cite{kobyzev2020normalizing}, Generative Adversarial Networks (GAN) \cite{goodfellow2014generative}, 
entropic regularization and Sinkhorn Distances \cite{cuturi2013sinkhorn,COTFNT}, Fisher information  regularization \cite{OT_Fisher_2018}, our method differs either in training problem formulation or in imposing fewer  constraints on the finite data samples. For example, our mapping is not required to be invertible as in NF, the training objective is not min-max as in GAN for image generation, there is no regularization effect such as blurring or noise. We believe that our method is better tailored to our problem of invariant measure learning by using a moderate amount of training data generated from the IPM. Detailed comparison will be left for future work. 

%deviates from classical OT study (like entropy \cite{cuturi2013sinkhorn} or embedding [Approximate earth mover’s distance in linear time
%Sameer Shirdhonkar and David W. Jacobs]) as it does not ask for finite state space (called discrete EMD with histogram in some literature);  Neural Ordinary Differential Equations \cite{chen2018neural}; Wasserstein GAN \cite{arjovsky2017wasserstein}}

\medskip
The rest of the paper is organized as follows. In Section 2, we review the basic idea of DNNs and their properties, as well as Wasserstein distance. In Section 3, we introduce our DeepParticle method for learning and predicting invariant measures based on the 2-Wasserstein distance minimization. In Section 4, we present numerical results to demonstrate the performance of our method. Finally, concluding remarks are made in Section 5.
	
\section{Preliminaries}
\subsection{Artificial Neural Network}
\noindent
In this section, we introduce the definition and approximation properties of DNNs. There are two ingredients in defining a DNN. The first one is a (vector) linear function of the form $T:\mathbb{R}^n\rightarrow \mathbb{R}^m$, defined as $T(x)=Ax+b$, where $A=(a_{ij})\in \mathbb{R}^{m\times n}$, $x\in R^{n}$ and $b\in \mathbb{R}^m$. The second one is a nonlinear activation function $\sigma:\mathbb{R}\rightarrow \mathbb{R}$. A frequently used activation function, known as the rectified linear unit (ReLU), is defined as $\sigma(x)=\max(0,x)$ \cite{lecun2015deep}. In the neural network literature, the sigmoid function is another frequently used activation function, which is defined as $\sigma(x)=(1+e^{-x})^{-1}$. By applying the activation function in an element-wise manner, one can define (vector) activation function $\sigma: \mathbb{R}^m\rightarrow \mathbb{R}^m$.
	
Equipped with those definitions, we are able to define a continuous function $F(x)$ by a composition of linear transforms and activation functions, i.e., 
	\begin{equation}\label{eqn:eg3layernet}
		F(x)=T^{k}\circ\sigma\circ T^{k-1}\circ\sigma
		\cdot\cdot\cdot\circ T^{1}\circ\sigma\circ T^{0}(x),
	\end{equation}
	where $T^{i}(x)=A_ix+b_i$ with $A_i$ be undetermined matrices and $b_i$ be undetermined vectors, and $\sigma(\cdot)$ is the element-wise defined activation function. Dimensions of $A_i$ and $b_i$ are chosen to make \eqref{eqn:eg3layernet} meaningful. Such a DNN is called a $(k+1)$-layer DNN, which has $k$ hidden layers. Denoting all the undetermined coefficients (e.g., $A_i$ and $b_i$) in \eqref{eqn:eg3layernet} as $\theta\in\Theta$, where $\theta$ is a high-dimensional vector and $\Theta$ is the space of $\theta$. The DNN representation of a continuous function can be viewed as	
	
	\begin{align}\label{eqn:solution_DNN}
		F=F(x;\theta).
	\end{align}
	Let $\mathbb{F}=\{ F(\cdot,\theta)|\theta\in\Theta\}$ denote the set of all expressible functions by the DNN parameterized by $\theta\in\Theta$. Then, $\mathbb{F}$ provides an efficient way to represent unknown continuous functions, in contrast  with a linear solution space in classical numerical methods, e.g. the space of linear nodal basis functions in the finite element methods and orthogonal polynomials in the spectral methods. 

\subsection{Wasserstein distance and optimal transportation}
\noindent	
Wasserstein distances are metrics on probability distributions inspired by the problem of optimal mass transportation. They measure the minimal effort required to reconfigure the probability mass of one distribution in order to recover the other distribution. They are ubiquitous in mathematics, especially in fluid mechanics, PDEs, optimal transportation, and probability theory \cite{villani2021topics}.
	
One can define the $p$-Wasserstein distance between probability measures $\mu$ and $\nu$ on a metric space $Y$ with distance function $dist$ by 

	\begin{align}\label{def:p-Wdistance}
	W_{p}(\mu, \nu):=\left(\inf _{\gamma \in \Gamma(\mu, \nu)} \int_{Y \times Y} dist (y', y)^{p} \mathrm{~d} \gamma(y', y)\right)^{1 / p}
	\end{align}
	where $\Gamma(\mu, \nu)$ is the set of probability measures $\gamma$ on $Y\times Y$ satisfying $\gamma(A\times Y)=\mu(A)$ and $\gamma(Y\times B)=\nu(B)$ for all Borel subsets $A,B \subset Y$. Elements $\gamma \in \Gamma(\mu, \nu)$ are called couplings of the measures $\mu$ and $\nu$, i.e., joint distributions on $Y\times Y$ with marginals $\mu$ and $\nu$ on each axis. 
	
	In the discrete case, the definition \eqref{def:p-Wdistance} has a simple intuitive interpretation: given a $\gamma \in \Gamma(\mu, \nu)$ and any pair of locations $(y',y)$, the value of $\gamma(y',y)$ tells us what proportion of $\mu$ mass at $y'$ should be transferred to $y$, in order to reconfigure $\mu$ into $\nu$. Computing the effort of moving a unit of mass from $y'$ to $y$ by $dist(y',y)^{p}$ yields the interpretation of $W_{p}(\mu, \nu)$ as the minimal effort required to reconfigure $\mu$ mass distribution into that of $\nu$. 
	A cost function $c(y',y)$ on $Y\times Y$ tells us how much it costs to transport one unit of mass from location $y'$ to location $y$. When the cost function $c(y',y)=dist(y',y)^{p}$, the $p$-Wasserstein distance \eqref{def:p-Wdistance} reveals the Monge-Kantorovich optimization problem, see \cite{villani2021topics} for more exposition. 
	
	In a practical setting \cite{COTFNT}, the closed-form solution of $\mu$ and $\nu$ may be unknown, instead only $N$ independent and identically distributed (i.i.d.) samples of $\mu$ and $\nu$ are available. We approximate the probability measures $\mu$ and $\nu$ by empirical distribution functions:
	
	\begin{align}\label{def:empiricalPDF}
	\mu=\frac{1}{N}\sum_{i=1}^{N}\delta_{x_i} \quad \text{and}\quad  
	\nu=\frac{1}{N}\sum_{j=1}^{N}\delta_{y_j}.
	\end{align}
	Any element in $\Gamma(\mu,\nu)$ can clearly be represented by an $N\times N$ doubly stochastic matrix \cite{sinkhorn1964relationship}, denoted as transition matrix,  $\gamma=(\gamma_{ij})_{i,j}$ satisfying:   
	
	\begin{align}
	\label{def:bistochasticity}
		\gamma_{ij} \geq 0; \quad \quad
	 \forall j, ~\sum_{i=1}^{N}\gamma_{ij}=1;
	 \quad \quad  
	 \forall i, ~\sum_{j=1}^{N}\gamma_{ij}=1.
	\end{align}	
The empirical distribution functions allow us to approximate different measures. And the doubly stochastic matrix provides a practical tool to study the transformation of measures via minimizing the Wasserstein distance between different empirical distribution functions. Notice that when the number of particles becomes large, it is expensive to find the transition matrix $\gamma$ to calculate discrete Wasserstein distance. We propose an efficient submatrix sampling method to overcome this difficulty in the next section. 

\section{Methodology}
\subsection{Physical parameter dependent neural networks}\label{subsec:parNN}
\begin{figure}[ht]
		\centering
	
\begin{tikzpicture}[
	ionode/.style={circle, draw=green!60, fill=green!5, very thick, minimum size=2mm},
	sqnode/.style={rectangle, draw=brown!60, fill=red!5, very thick, minimum size=5mm},
	layernode/.style={ellipse, draw=blue!60, fill=red!5, very thick, aspect=0.7, minimum size=10mm},
	wblayernode/.style={rounded rectangle, draw=cyan!60, fill=red!5, very thick, aspect=0.7, minimum size=5mm},
	node distance=4mm,
	]
	%Nodes
	\node[ionode]  (xlayer) {x};
	\node[layernode]  (l1) [below=of xlayer]  {layer1};
	\node[layernode]   (l1-2)   [right=of l1] {layer1\_2};
		\node[ionode]  (parlayer) [above=of l1-2] {$\eta$};
	\node[layernode]   (l2-2)   [below=of l1-2] {layer2\_2};
	\node[layernode]   (l3-2)   [below=of l2-2] {layer3\_2};
	\node[sqnode]  (wb) [right=of l2-2]  {weight and bias};
			\node[wblayernode] (wbl3) [above=of wb] {par-net: layer3};
	\node[wblayernode] (wbl2) [above=of wbl3] {par-net: layer2};
		\node[wblayernode] (wbl1) [above=of wbl2] {par-net: layer1};
	\node[layernode]  (l2) [below=of l1]  {layer2};
	\node[layernode]  (l3) [below=of l2]  {layer3};
	\node[layernode]  (l4) [below=of l3]  {layer4};
	\node[layernode]  (l5) [below=of l4]  {layer5};
	\node (mlayer) [below=of l5] {\Huge ...};
	\node[layernode]  (l11) [below=of mlayer]  {layer11};
\node[layernode]  (l12) [below=of l11]  {layer12};
\node[ionode]    (output)  [below=of l12] {$f(x)$};
	
	%Lines
	\draw[->] (xlayer.south) -- (l1.north);
	\draw[->] (xlayer.south) .. controls +(down:2mm) and +(up:2mm) .. (l1-2.north);
		\draw[->] (l1.south) .. controls +(down:2mm) and +(up:2mm) .. (l2-2.north);
				\draw[->] (l2.south) .. controls +(down:2mm) and +(up:2mm) .. (l3-2.north);
	\draw[->] (l1-2.south) .. controls +(down:2mm) and +(up:2mm) .. (l2.north);
	\draw[->] (l2-2.south) .. controls +(down:2mm) and +(up:2mm) .. (l3.north);
	\draw[->] (l3-2.south) .. controls +(down:2mm) and +(up:2mm) .. (l4.north);
	\draw[->] (parlayer.south) .. controls +(down:2mm) and +(up:2mm) .. (l1.north);
	\draw[->] (parlayer.east) .. controls +(right:2mm) and +(left:2mm) .. (wbl1.west);
	\draw[-latex] (wb.west)  -- (l1-2.east);
		\draw[-latex] (wb.west)  -- (l2-2.east);
			\draw[-latex] (wb.west)  -- (l3-2.east);
				\draw[->] (wbl1.south)--(wbl2.north);
					\draw[->] (wbl2.south)--(wbl3.north);
						\draw[->] (wbl3.south)--(wb.north) node[midway,right] {no activation};
	\draw[->] (l1.south)--(l2.north);
	\draw[->] (l2.south)--(l3.north);
	\draw[->] (l3.south)--(l4.north);
	\draw[->] (l4.south)--(l5.north);
	\draw[->] (l2.west)  .. controls +(left:7mm) and +(left:7mm) .. (l4.west) node[midway,left] {skip};
	\draw[->] (l3.west)  .. controls +(left:7mm) and +(left:7mm) .. (l5.west) node[midway,left] {skip};
	
	\draw[dashed,->] (l4.west)  .. controls +(left:7mm) and +(left:12mm) .. (mlayer.west) node[midway,left] {skip};
	\draw[dashed,->] (mlayer.west)  .. controls +(left:12mm) and +(left:7mm) .. (l12.west) node[midway,left] {skip};
	\draw[->] (mlayer.south)--(l11.north);
	\draw[->] (l11.south)--(l12.north);
	\draw[->] (l12.south)--(output.north) node[midway,right] {linear, no activation};
	\draw[->](xlayer.west).. controls +(left:30mm) and +(left:30mm) ..(output.west) node[midway,left] {linear, no activation};
	\draw[->] (l5.south)--(mlayer.north);

\end{tikzpicture}
\caption{Layout of the proposed deep network where $\eta$ is physical parameter input for learning its implicit dependence in the output $f(x)$.}\label{NetworkLayout}
\end{figure}
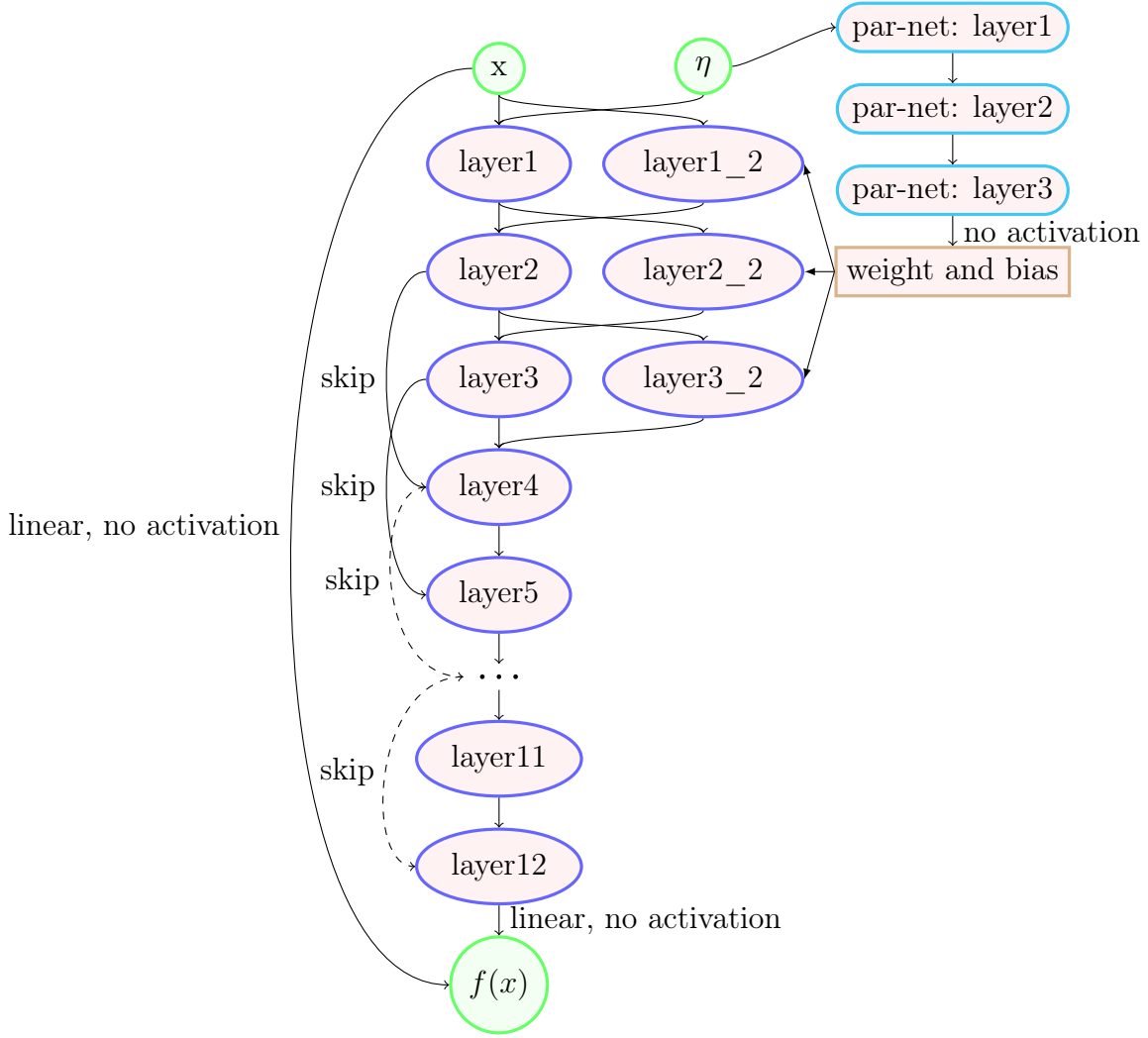

	%plan: compare result direct include $\sigma$ and shallow $\sigma$ network that generate weight and bias of the main network. result : indeed work better!
\noindent	Compared with general neural networks, we propose a new architecture to learn the dependency of physical parameters. Such type of network is expected to have independently batched input, and the output should also rely on some physical parameter that is shared by the batched input. Thus, in addition to concatenating the physical parameters as input, we also include some linear layers whose weights and biases are generated from a shallow network; see Fig. \ref{NetworkLayout} for the layout of the proposed network used in later numerical experiments. 
%	\medskip
	
To be more precise, we let the network take on two kinds of input, $X$ and $\eta$. Their batch sizes are denoted as $N$ and $n_{\eta}$ respectively. Then, there are $n_{\eta}$ sets of $X$, such that $X$ in each set shares the same physical parameter $\eta$. From $\mathtt{layer1}$ to $\mathtt{layer12}$ are general linear layers in which the weights and biases are randomly initialized and updated by Adams descent during training. They are $20$ in width and Sigmoid function is applied as activation between adjacent layers. Arrows denote the direction in forward propagation. From $\mathtt{layer1\_2}$ to $\mathtt{layer3\_2}$ are linear layers with $20$ in width, but every element of weight matrix and bias vector is individually generated from the $\mathtt{par-net}$ specified on the right. Input of $\mathtt{par-net}$ is the shared parameter, namely $\eta$. From $\mathtt{par-net:layer1}$ to $\mathtt{par-net:layer3}$ are general linear layers with $10$ in width. For example, given the physical parameter $\eta$ is $d_\eta$ dimension, to generate a $20\times20$ weight matrix, we first tile $\eta$ to $\mathtt{bsz}\times20 \times 20 \times d_\eta$ tensor ${\boldsymbol\eta}$ (boldface for tensor). Here $\mathtt{bsz}$ denotes the shape of input batches which are independently forwarded in the network. In our numerical examples,  $\mathtt{bsz}=n_{\eta}\times N$. Then, we introduce a $20\times  d_\eta \times 10$ tensor as weight matrix $w$ in $\mathtt{par-net:layer1}$ and do matrix multiplication of ${\boldsymbol\eta}$ and  $w$ on the last two dimensions while keeping dimensions in front.  The third dimension in $w$ is the width of the $\mathtt{par-net:layer3}$. Shapes of weight tensor in $\mathtt{par-net:layer2}$ and $\mathtt{par-net:layer3}$ are $20 \times 10 \times 10$ and $20\times 10 \times 1$ respectively. The weight parameters of the  $\mathtt{par-net}$ are  randomly initialized and updated by gradient descent during training. We also include some skip connections as in $\mathtt{Resnet}$ to improve the performance of deeper layers in our network. In addition to standard $\mathtt{Resnet}$ short cuts, we design a linear propagation path directly from the input $x$ to the output $f(x)$. In numerical experiment shown later, such connection helps  
	%preserving local perturbation of the input data. It will then 
	avoid output being over-clustered. There is no activation function between the last layer and output.

\subsection{DeepParticle Algorithms}
\noindent	Given distributions $\mu$ and $\nu$ defined on metric spaces $X$ and $Y$, we aim to construct a transport map $f^0:X \to Y$ such that
%	\begin{align}
		$f^{0}_{*}(\mu)= \nu$, where star denotes the push forward of the map. 
%	\end{align}
On the other hand, given function $f:X \to Y$, the $p$-Wasserstein distance between  $f_*(\mu)$ and $\nu$ is defined by:

\begin{align}\label{def:Wdistance}
	W_{p}(f_*(\mu), \nu):=\left(\inf _{\gamma \in \Gamma(f_*(\mu), \nu)} \int_{Y \times Y} dist(y', y)^{p} \mathrm{~d} \gamma(y', y)\right)^{1 / p},
\end{align}
where $\Gamma(f_*(\mu), \nu)$ denotes the collection of all measures on $Y \times Y$ with marginals $f_*(\mu)$ and $\nu$ on the first and second factors respectively and $dist$ denotes the metric (distance) on $Y$. A 
straightforward derivation yields:

\begin{align}\label{def:Wdistance1}
    W_{p}(f_*(\mu), \nu)=\left(\inf _{\gamma \in \Gamma(\mu, \nu)} \int_{X \times Y} dist(f(x), y)^{p} \mathrm{~d} \gamma(x, y)\right)^{1 / p},
\end{align}
where $\Gamma(\mu, \nu)$ denotes the collection of all measures on $X \times Y$ with marginals $\mu$ and $\nu$ on the first and second factors respectively and still $dist$ denotes the metric (distance) on $Y$.

\paragraph{Network Training Objective}
Our DeepParticle algorithm does not assume the knowledge of closed form distribution of $\mu$ and $\nu$, instead we have i.i.d. samples of $\mu$ and $\nu$ namely, $x_i$ and $y_j$, $i,j=1,\cdots,N$, as training data. Then a discretization of \eqref{def:Wdistance1} is:

\begin{align}\label{def:Wdistance_approx}
\hat{W}(f):=\left(\inf_{\gamma \in \Gamma^N}\sum_{i,j=1}^N\, dist(f(x_i),y_j)^p \gamma_{ij}\right)^{1/p},
\end{align}
where $\Gamma^N$ denotes all $N\times N$ doubly stochastic matrices.

Let the map (represented by neural network in Fig. \ref{NetworkLayout}) be $f_\theta(x;\eta)$ where $x$ is the input, $\eta$ is the shared physical parameter and $\theta$ denotes all the trainable  parameters in the network. Clearly $\hat{W}(f_\theta)\geq 0$. In case of  $X=Y=\mathbb{R}^d$ equipped with Euclidean metric, we take $p=2$. The training loss function is

\begin{align}\label{def:penalty-1}
	\hat{W}^2(f_\theta):=\sum_{r=1}^{n_{\eta}}\left(\inf_{\gamma_r \in \Gamma^N}\sum_{i,j=1}^N|f_\theta(x_{i,r};\kappa_r)-y_{j,r}|^2\gamma_{ij,r}\right).
\end{align}

\paragraph{Iterative method in finding transition matrix $\gamma$}
 To minimize the loss function \eqref{def:penalty-1}, we update parameters $\theta$ of $f_\theta$ with the classical Adams stochastic gradient descent, and  alternate with updating the transition matrix $\gamma$. %Since the processes of finding $\gamma$ under different input of physical parameter $\eta$ are mutually independent, we ignore the summation of $r=1,\cdots, n_{\eta}$ in Eq. \eqref{def:penalty-1} for simplicity in later derivation.
 \medskip
 
 We now present a {\it mini-batch linear programming algorithm} to find the best $\gamma$ for each  %$\|f_\theta(x_i)-y_j\|^2$ 
 inner sum of \eqref{def:penalty-1}, while suppressing $k_r$ dependence in $f_\theta$. 
 Notice that the problem \eqref{def:penalty-1} is a linear program on the bounded convex set $\Gamma^N$ of vector space of real $N\times N$ matrices. By Choquet's theorem, this problem admits solutions that are extremal points of $\Gamma^N$. Set of all doubly stochastic matrix $\Gamma^N$ can be referred to as Birkhoff polytope. The Birkhoff–von Neumann theorem \cite{schrijver2003combinatorial} states that such polytope is the convex hull of all permutation matrices, i.e., those matrices such that $\gamma_{ij}=\delta_{j,\pi(i)}$ for some permutation $\pi$ of $\{1,...,N\}$, where $\delta_{jk}$ is the Kronecker symbol. 
 
 The algorithm is defined iteratively. In each iteration, we select columns and rows and solve a linear programming sub-problem under the constraint that maintains column-wise and row-wise sums of the corresponding sub-matrix of $\gamma$. To be precise, let $\{i_k\}_{k=1}^M$, $\{j_l\}_{l=1}^M$ ($M\ll  N$) denote the index chosen from $\{1,2, \cdots, N\}$ without replacement. The cost function of the sub-problem is 
 
\begin{align}\label{def:lp-sub-cost}
	C(\gamma^*):=\sum_{k,l=1}^M|f_\theta(x_{i_k})-y_{j_l}|^2\gamma^*_{i_kj_l}
\end{align}
subject to

\begin{align}\label{def:lp-sub-cons}
\left\{	\begin{array}{l}
		\sum_{k=1}^M \gamma^*_{i_k,j_l}=	\sum_{k=1}^M {\gamma}_{i_k,j_l}\quad\forall l=1,\cdots,M\\
			\sum_{l=1}^M \gamma^*_{i_k,j_l}=	\sum_{l=1}^M {\gamma}_{i_k,j_l}\quad\forall \, k=1,\cdots,M\\
	\gamma^*_{i_kj_l}\geq 0 \quad \forall \, k,l = 1,\cdots,M,
\end{array}\right.
\end{align}
where ${\gamma}_{i_k,j_l}$ are from the previous step. This way, the column and row sums of $\gamma$ are preserved by the update.  The linear programming sub-problem of much smaller size is solved by the interior point method \cite{wright1997primal}.
\medskip

We observe that the global minimum of $\gamma$ in  \eqref{def:penalty-1} is also the solution of sub-problems \eqref{def:lp-sub-cost} with arbitrarily selected rows and columns, subject to the row and column partial sum values of the global minimum. The selection of rows and columns can be one's own choice. In our approach, in each step after gradient descent, apart from random sampling of rows and columns, we also perform a number of searches for rows or columns with the largest entries to speed up computation; see Alg. \ref{alg:pick_col}. 
\bigskip

 \begin{algorithm}[H]
	\SetAlgoLined
	\KwResult{Given transition matrix $\gamma$, randomly search $M$ rows/columns with largest elements (``pivots'').}
	Randomly pick $i_1$ as the first row;\\
	find $j_1$ such that $\gamma_{i_1j_1}$ is the largest among $\{\gamma_{i_11},\cdots,\gamma_{i_1N}\}$;\\
	\For{$k\gets2$ \KwTo $M$ }{
		find $i_k$ such that $\gamma_{i_kj_{k-1}}$ is the largest among $\{\gamma_{i'j_{k-1}}\}_{i'\in\{1,\cdots, N\}\setminus\{i_1,\cdots ,i_{k-1}\}}$;\\
		find $j_k$ such that $\gamma_{i_kj_k}$ is the largest among $\{\gamma_{i_kj'}\}_{j'\in\{1,\cdots, N\}\setminus\{j_1,\cdots ,j_{k-1}\}}$;
}
	\caption{Random Pivot Search}\label{alg:pick_col}
\end{algorithm}
\medskip

The cost of finding optimal $\gamma$ increases as $N$ increases, however, the network itself is independent of $\gamma$. After training, our network acts as a sampler from  some target distribution $\nu$ 
without assumption of closed-form distribution of $\nu$. At this stage, the input data is no longer limited by  training data, an arbitrarily large amount of samples approximately obeying $\nu$ can be generated through $\mu$ (uniform distribution).

\paragraph{Update of training data}
  Note that given any fixed set of $\{x_i\}$ and $\{y_j\}$ (training data), we have developed the iterative method to calculate the optimal transition matrix $\gamma$ in \eqref{def:penalty-1} and update network parameter $\theta$ at the same time. However, in more complicated cases, more than one set of data (size of which is denoted as $\mathtt{bsz}$) should be assimilated. The total number of data set is denoted $N_{dict}$. For the second and later set of training data, the network is supposed to establish some accuracy. So before updating the network parameter, we first utilize our iterative method to find an approximated new $\gamma$ for the new data feed to reach a preset level measured by the  {\it normalized Frobenius norm} $\|\cdot \|_{fro}$ of $\gamma$ ( $\|\gamma \|_F$ divided by the square root of its row number).  
   The full training process is outlined in Alg. \ref{alg:1}.

\begin{algorithm}[htbp]
	\SetAlgoLined
		Randomly initialize weight parameters  $\theta$ in network $f_\theta:\mathbb{R}^d\to \mathbb{R}^d$; \\
	\Repeat{given number of training mini-batches, $N_{dict}$}{
	\For{ physical parameter set $r\gets0$ \KwTo $n_{\eta}$ }{
	randomly select $\{x_{i,r}\}$, $\{y_{j,r}\}$, $i,j=1:N$ from i.i.d. samples of input and target distribution with respect to physical parameter $\eta_r$;\\
	$\gamma_{ij,r}=1/N$;\\
	}
	\If{not the first training mini-batch}{
		\For{ physical parameter set $r\gets0$ \KwTo $n_{\eta}$ }{
		$P_r=\sum_{i,j=1}^N|f_\theta(x_{i,r},\eta_r)-y_{j,r}|^2\gamma_{ij,r}$;\\
	\While{$||\gamma_r||_{fro}<tol$}{
		randomly (or by Alg.\ref{alg:pick_col}) choose $\{i_{k,r}\}_{k=1}^M$, $\{j_{l,r}\}_{l=1}^M$ from $\{1,2, \cdots, N\}$ without replacement; \\
		solve the linear programming 
		sub-problem \eqref{def:lp-sub-cost}-\eqref{def:lp-sub-cons} 
		to get $\gamma_r^*$;\\
		update $\{\gamma_{i_{k,r}j_{l,r}}\}_{k,l=1}^M$ with $\{\gamma^*_{i_{k,r}j_{l,r}}\}_{k,l=1}^M$.
	}
	}
	}
	\Repeat{given steps for each training mini-batch}{
		$P=\sum_{r=1}^{N_r}\sum_{i,j=1}^N|f_\theta(x_{i,r},\eta_r)-y_{j,r}|^2\gamma_{ij,r}$;\\
		$\theta\leftarrow \theta - \delta_1\nabla_{\theta} P$, $\delta_1$ is the learning step size;\\
		\Repeat{given linear programming steps, $N_{LP}$}{
			\For{ physical parameter set $r\gets0$ \KwTo $n_{\eta}$ }{
					$P_r=\sum_{i,j=1}^N|f_\theta(x_{i,r},\eta_r)-y_{j,r}|^2\gamma_{ij,r}$;\\
			randomly (or by Alg.\ref{alg:pick_col}) choose $\{i_{k,r}\}_{k=1}^M$, $\{j_{l,r}\}_{l=1}^M$ from $\{1,2, \cdots, N\}$ without replacement; \\
		solve the linear programming 
		sub-problem \eqref{def:lp-sub-cost}-\eqref{def:lp-sub-cons} 
		to get $\gamma_r^*$;\\
		update $\{\gamma_{i_{k,r}j_{l,r}}\}_{k,l=1}^M$ with $\{\gamma^*_{i_{k,r}j_{l,r}}\}_{k,l=1}^M$.
	}
	}
	}
	} 
	$\mathbf{Return}$
	\caption{DeepParticle Learning}\label{alg:1}
\end{algorithm}

\section{Numerical Examples}
\subsection{Mapping uniform to normal distribution}
\noindent
For illustration, we first apply our algorithm to learn a map from 1D uniform distribution on $[0,1]$ to 1D normal distribution with zero mean  and various standard derivation $\sigma_1$. The $\sigma_1$ refers to one dimensional $\eta$ in the network of Fig.\ref{NetworkLayout}. As training data, we independently generate $n_{\sigma_1}=8$ sets of $N=1500$ uniformly distributed points, also $N=1500$ normally distributed sample points with zero mean and standard derivation $\sigma_1$ equally spaced in $[2,3.75]$. During training, we aim to find a $\sigma_1$-dependent network such that the output from the $8$ input data sets along with their $\sigma_1$ values approximate normal distributions with standard derivation $\sigma_1$, respectively. The layout of network is shown in Fig. \ref{NetworkLayout}.  The total number of training steps is $10^4$ with initial learning rate $0.02$ and  $N_{dict}=1$ data set. After each update of parameters, we solve the optimization problem of transition matrix $\gamma$, i.e. Eq.\eqref{def:lp-sub-cost} under Eq.\eqref{def:lp-sub-cons}, with selection of rows and columns by Alg. \ref{alg:pick_col} for $N_{LP}=5$ times. After training, we generate $N=40000$ uniformly distributed test data points and apply the networks with $\sigma_1=2$, $3$, $4$, $5$. 

\begin{figure}[htbp]
	\centering
	\subfigure[$\sigma_1=2$]{\includegraphics[width=0.34\linewidth]{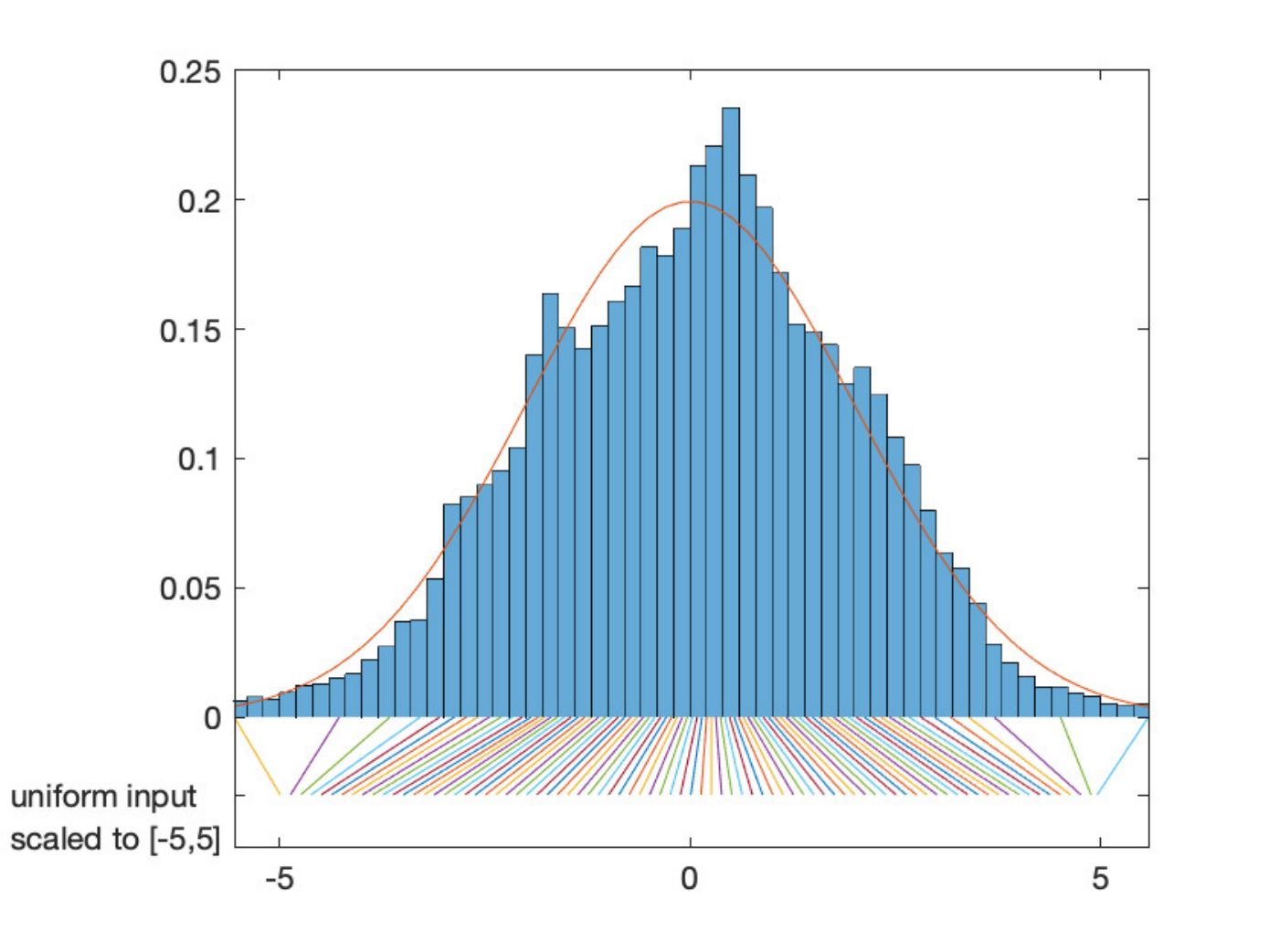}}
	\subfigure[$\sigma_1=3$]{\includegraphics[width=0.34\linewidth]{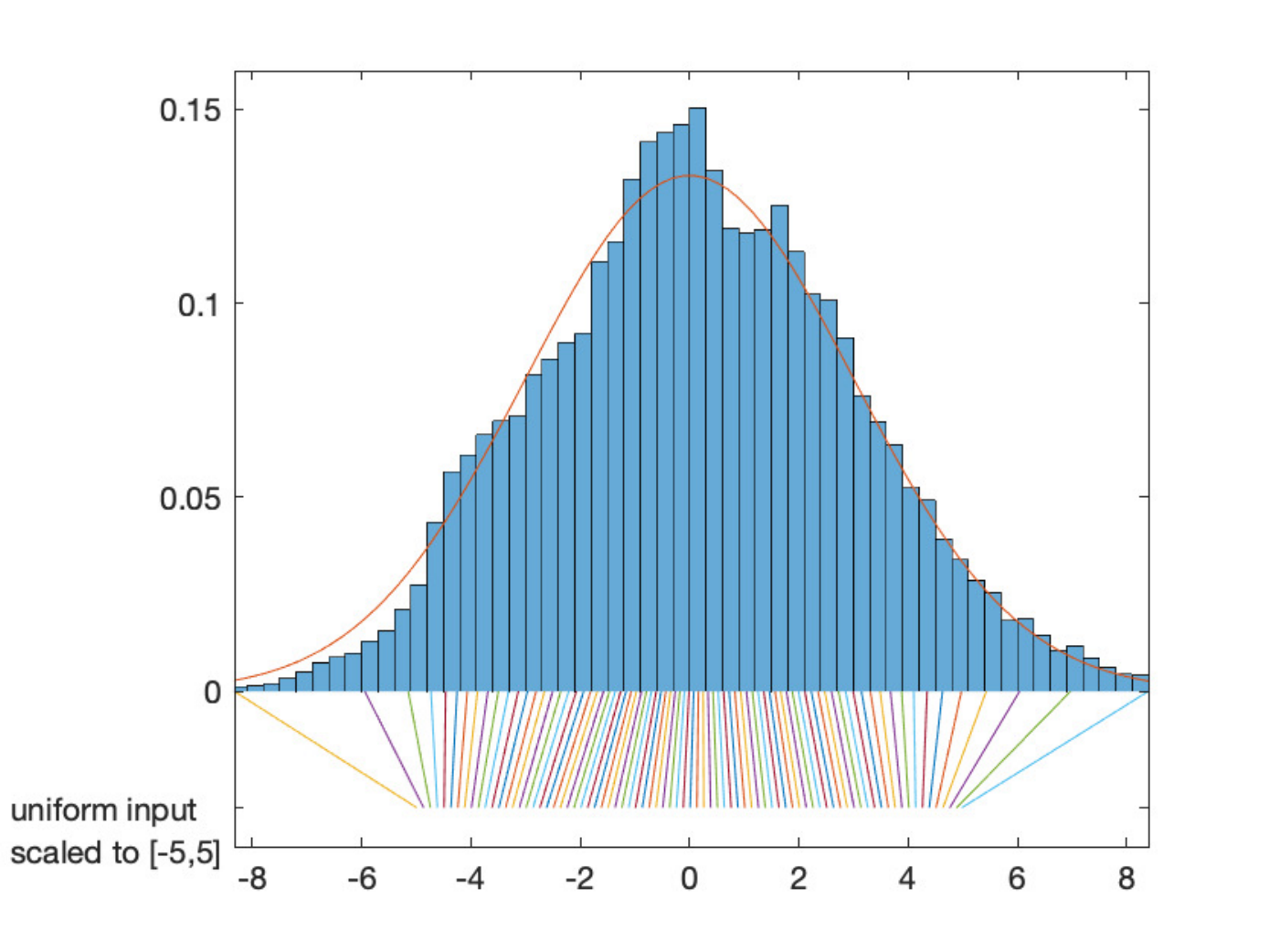}}
	\\
	\subfigure[$\sigma_1=4$]{\includegraphics[width=0.34\linewidth]{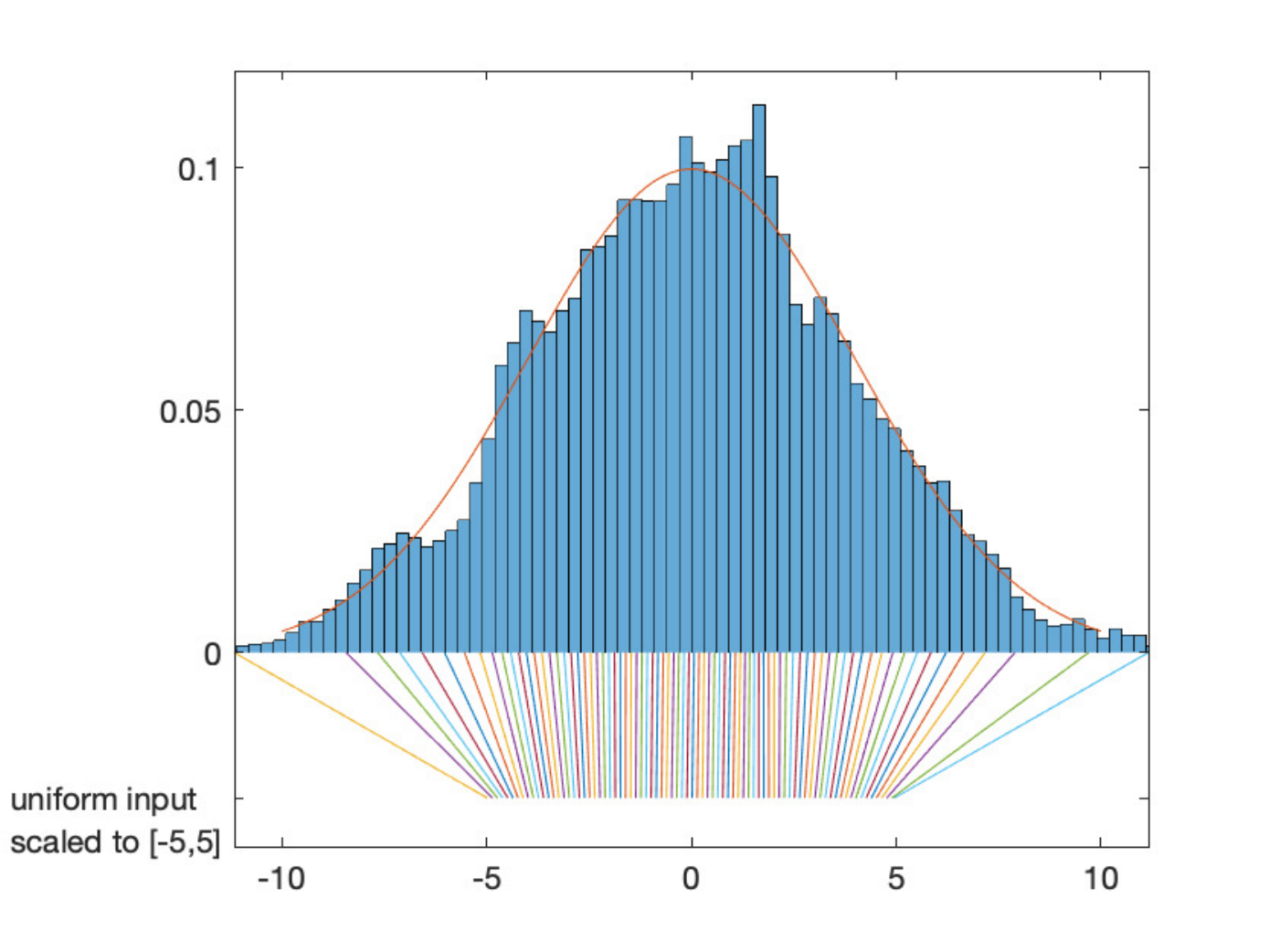}}
	\subfigure[$\sigma_1=5$]{\includegraphics[width=0.34\linewidth]{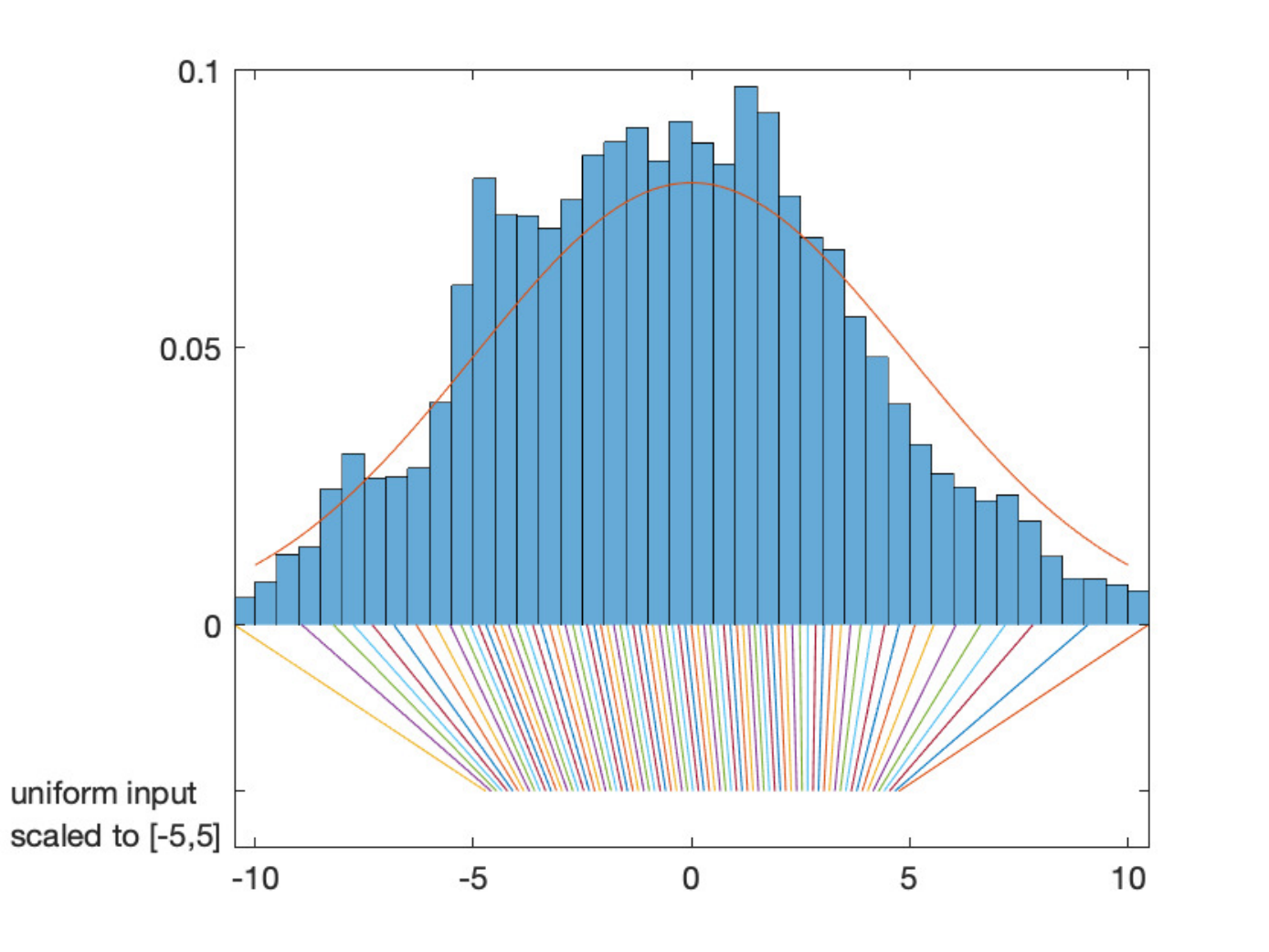}}
	\caption{Generated histograms with the red lines being the ground truth pdfs. {Lines under the pdfs connect inputs, scaled to $Unif[-5,5]$, and their corresponding outputs of the network.}}
	\label{eg1:uniform2normal}
\end{figure}

In Fig. \ref{eg1:uniform2normal},  we plot the output histograms. Clearly, the empirical distributions vary with $\sigma_1$ and fit the reference normal probability distribution functions (pdfs) in red lines. We would like to clarify that even though the dependence of $\sigma_1$ in this example is `linear', the linear propagation directly from input to output does not depend on $\sigma_1$ and output $\mathtt{par-net}$ is no longer a linear function of $\sigma_1$. {In addition, we plot the input-output pairs at the bottom of each empirical distribution. Each line connects the input (for better visualization we scale it to $Unif[-5,5]$) and its corresponding output from the network. In one dimensional case, it is well known that the inverse 
of the cumulative distribution function is the optimal transport map. It is a \emph{monotone} function that maps uniform distribution to the quantile of the target distribution. In the plotted input-output pair, we can see there are no lines crossing. It verifies that our network finds the \emph{monotone} transport map.} Our experiments here and below are all carried out on a quad-core CPU desktop with an RTX2080 8GB GPU at UC Irvine. 

\subsection{Computing front speeds in complex fluid flows}
\noindent
Front propagation in complex fluid flows arises in many scientific areas such as turbulent combustion, chemical kinetics,  biology, transport in porous media, and industrial deposition processes \cite{xin2009}. A fundamental problem is to analyze and compute large-scale front speeds. An extensively studied model problem is the reaction-diffusion-advection (RDA) equation with Kolmogorov-Petrovsky-Piskunov (KPP) nonlinearity \cite{kolmogorov1937}: 

\begin{align}
u_t  = \kappa\, \Delta_{\textbf{x}} u + (\textbf{v} \cdot \nabla_{\textbf{x}})\, u + u(1-u), \quad  t\in \mathbb{R}^{+}, \quad \textbf{x}=(x_1,...,x_d)^{T}\in \mathbb{R}^{d},  \label{KPP-eq}
\end{align}
where $\kappa$ is diffusion constant,  
$\textbf{v}$ is an incompressible velocity field (precise definition given later), and $u$ is the concentration of reactant. 
Let us consider velocity fields $\textbf{v}=\textbf{v}(t,\textbf{x})$ to be  $T$-periodic in space and time, which 
contain the celebrated Arnold-Beltrami-Childress (ABC) and Kolmogorov flows as well as their variants with chaotic streamlines \cite{CG95,KLX_21}. 
The solutions from compact non-negative initial data 
spread along direction $\textbf{e}$ 
with speed \cite{nolen2005existence}:
%\[ c^{*}(\textbf{e})=\inf_{\lambda>0}\; \mu(\lambda)/\lambda, \] 
\[ c^{*}(\textbf{e})=\inf_{\alpha>0}\; \lambda(\alpha)/\alpha, \] 
where $\lambda(\alpha)$ is the principal eigenvalue of 
the parabolic operator $\partial_t-\mathcal{A}$ with:
\begin{equation} \label{KPPTimePeriodic}
\mathcal{A}w:=\kappa\, \Delta_{\textbf{x}} w + (2\alpha\, \textbf{e}+\textbf{v})\cdot
\nabla_{\textbf{x}}w + \big(\kappa \,  \alpha^2+\alpha\, \textbf{v}\cdot \textbf{e} + 1 \big)w, 
\end{equation} 
on the space domain $\mathbb{T}^{d}:=[0, T]^{d}$ (periodic boundary condition).
It is known \cite{nolen2005existence} that $\lambda(\alpha)$ is  
convex in $\alpha$, 
and superlinear for 
large $\alpha$.
The operator $\mathcal{A}$ in \eqref{KPPTimePeriodic} is a sum  $\mathcal{A}=\mathcal{L}+\mathcal{C}$, with
\begin{equation} 
\mathcal{L}:=\kappa\, \Delta_{\textbf{x}} \cdot  + (2\, \alpha\, \textbf{e}+\textbf{v})\cdot\nabla_{\textbf{x}}\cdot, \; \; 
\mathcal{C}:=c(t,\textbf{x})\cdot =\big(\kappa \, \alpha^2+\, \alpha\, \textbf{v}\cdot \textbf{e} + 1 \big)\cdot
\end{equation}
where $\mathcal{L}$ is the generator of a Markov process, 
and $\mathcal{C}$ acts as a potential. 
The Feynman-Kac (FK) formula  \cite{Fr_book} gives 
$\lambda(\alpha)$ a stochastic representation:
\begin{equation}
\lambda = \lim_{t \rightarrow \infty}\, t^{-1}\ln \left (\mathbb{E} \exp \left \{ \int_{0}^{t}\, c(t-s, \boldsymbol{X}^{t,\boldsymbol{x}}_{s})\, ds \right \}\right ). 
\label{FK}
\end{equation}
In Eq.\eqref{FK}, $\boldsymbol{X}^{t,\boldsymbol{x}}_{s}$ satisfies the following stochastic differential equation 
\begin{equation} \label{SDE}
 d\, \boldsymbol{X}^{t,\boldsymbol{x}}_{s} = 
\boldsymbol{b}(t-s,  \boldsymbol{X}^{t,\boldsymbol{x}}_{s})\, ds + 
\sqrt{2\kappa} \, d \boldsymbol{W}_{s}, \;  \boldsymbol{X}^{t,\boldsymbol{x}}_{0}=\boldsymbol{x}, 
\end{equation} 
where the drift term $\textbf{b}=2\, \alpha\, \textbf{e}+\textbf{v}$ 
and $\textbf{W}(s)$ is the standard $d$-dimensional Wiener process. The expectation $\mathbb{E}(\cdot)$ in (\ref{FK}) is over $\textbf{W}(t)$. Directly applying (\ref{FK}) and Monte Carlo method to compute $\lambda(\alpha)$ is unstable 
as the main contribution to $\mathbb{E}(\cdot)$ comes from sample paths that visit maximal or minimal points of the potential function $c$, resulting in inaccurate or even divergent results. A computationally feasible alternative is a scaled version, the FK semi-group:
\[ \Phi_{t}^{c}(\nu)(\phi) := \frac{ \mathbb{E} [\phi(\boldsymbol{X}^{t,\boldsymbol{x}}_{t})\, 
 \exp \{ \int_{0}^{t}\, c(t-s, \boldsymbol{X}^{t,\boldsymbol{x}}_{s})\, ds \}]
}{ \mathbb{E} [ 
 \exp \{ \int_{0}^{t}\, c(t-s, \boldsymbol{X}^{t,\boldsymbol{x}}_{s})\, ds\}]}:= 
\frac{P^{c}_{t}(\nu)(\phi)}{ P^{c}_{t}(\nu)(1)}. \]
Acting on any initial probability measure $\nu$, $\Phi_{n T}^{c}(\nu)$ converges weakly to an {\it invariant measure} $\nu_c$ (i.e. $\Phi_{T}^{c}(\nu_c) =\nu_c$) as $n \rightarrow \infty$, for any smooth test function $\phi$. 
Moreover,
\begin{equation}
P^{c}_{t}(\nu_c) = \exp\{\lambda \, t\}\,\nu_c  \; \; {\rm or}\;\; 
\lambda = t^{-1} \, \ln\,  \mathbb{E}_{\nu_c} [P^{c}_{t}(\nu_c)].
\end{equation}
Given a time discretization step $\Delta t$, an interacting particle method (IPM) proceeds to discretize $\boldsymbol{X}^{t,\boldsymbol{x}}_{s}$ as $\boldsymbol{X}^{\Delta t}_{i}$, ($i=1,\cdots, n\times m$, where $m=\frac{T}{\Delta t}$) by Euler's method, approximates the evolution of probability measure $\Phi_{t}^{c}(\nu)$ by a particle system with a re-sampling technique to reduce variance. Let 

\begin{align}\label{FKdisc}
 P_n^{c,\Delta t}(\nu)(\phi) := \mathbb{E}\left [ \phi(\boldsymbol{X}^{\Delta t}_{nm})\,  
\exp\left \{ \Delta t \, \sum_{i=1}^{m} c( (m-i)\Delta t, \boldsymbol{X}^{\Delta t}_{i+(n-1)m})\right\} \right ].
\end{align}
 Then sampled FK semi-group actions on $\nu$:
\[ \Phi_{n}^{c,\Delta t}(\nu)(\phi)
=   \frac{P_{n}^{c,\Delta t}(\nu)(\phi) }{ P_{n}^{c,\Delta t}(\nu)(1)} \longrightarrow 
\int_{\mathbb{T}^{d}}\, \phi\, d\, \nu_{c,\Delta t},  \; \; {\rm as}\; n \uparrow \infty,\; \forall \; {\rm smooth}\; \phi, \]
where  $\nu_{c,\Delta t}$ is an invariant measure of the discrete map  $\Phi_{1}^{c,\Delta t}$, thanks to $\mathbf{b}$ being $T$-periodic in time.
There exists $q\in (0,1)$ so that \cite{IPM_2021}:
\begin{equation}
    \lambda_{\Delta t}^n:= (n T)^{-1}\, \ln  [P^{c,\Delta t}_{n}(\nu_{0})(1)]\xrightarrow[n\to\infty]{}T^{-1}\, \ln  [P^{c,\Delta t}_{1}(\nu_{c,\Delta t})(1)]
= \lambda + o((\Delta t)^q).
\end{equation}
 
The IPM algorithm below (Alg. \ref{gIPM1}) approximates the evolving measure as an empirical measure of a large number of genetic particles that undergo advection-diffusion ($\mathcal{L}$) and mutations ($\mathcal{C}$). The mutation relies on fitness and its normalization defined as in the FK semi-group. In Alg.\ref{gIPM1}, 
the evolution is phrased in  
$G$ generations, each moving and 
mutating $m$ times in a life span of $T$. The index $n$ in (\ref{FKdisc}) is changed to $G$.
\medskip

\begin{algorithm}[H]
	\SetAlgoLined
 Initialize first generation of $N_0$ particles 
$\boldsymbol{\xi}^{0}_{1}=(\xi_{1}^{0,1},\cdots,\xi_{1}^{0,N_0})$, 
uniformly distributed over $\mathbb{T}^d$ ($d\geq 2$).
Let $g$ be the generation number in approximating $\nu_{c,\Delta t}$.   
Each generation moves and replicates $m$-times, with a life span $T$ (time period), time step 
$\Delta t = T/m$.   

\While{$g=1:G-1$}{  

\While{$i=0:m-1$}{
 
$\boldsymbol{\zeta}^{i}_{g} 
\leftarrow$ one-step-advection-diffusion update on $\boldsymbol{\xi}^{i}_{g}$.
Define fitness $\boldsymbol{F}:= \exp\{ c(T- i\Delta t, \boldsymbol{\zeta}^{i}_{g})\, \Delta t \}$. 

$E_{g,i}:= \frac{1}{ \Delta t} \ln$ (mean population fitness).

 Normalize fitness to weight.
 $\boldsymbol{p}:= \boldsymbol{F}/SUM(\boldsymbol{F}).$ 

$\boldsymbol{\xi}^{i+1}_{g} \leftarrow $ resample 
$\boldsymbol{\zeta}^{i}_{g}$ via multinomial distribution with weight 
$\boldsymbol{p}$.}

$\boldsymbol{\xi}^{0}_{g+1} \leftarrow \boldsymbol{\xi}^{m}_{g}$, $E_{g} \leftarrow  {\rm mean}\,( E_{g,i})$ over $i$.
}

 Output: approximate $\lambda_{\Delta t}\leftarrow $ mean($E_{g}$), and 
$\boldsymbol{\xi}^{0}_{G}$.   
\caption{Genetic Interacting Particle Method}\label{gIPM1}
\end{algorithm}
\bigskip

Much progress has been made in the finite element computation 
\cite{shen2013finite2d,shen2013finite3d,JackXin:15} of the KPP principal eigenvalue problem  (\ref{KPPTimePeriodic}) particularly in steady 3D flows $\textbf{v}=\textbf{v}(\textbf{x})$. However, when $\kappa$ is small and the spatial dimension is 3, adaptive FEM can be extremely expensive. For 2D time periodic cellular flows, adaptive basis deep learning is found to improve the accuracy of reduced order modeling \cite{JackXinLyu:2017,JackXinLyu:2020}. 
Extension of deep basis learning to 3D in the Eulerian setting has not been attempted partly due to the costs of generating a sufficient amount of training data.

%\begin{algorithm}[H]
%	\SetAlgoLined
% Initialize first generation of $N_0$ particles 
%$\boldsymbol{\xi}^{0}_{1}=(\xi_{1}^{0,1},\cdots,\xi_{1}^{0,N_0})$, 
%uniformly distributed over $\mathbb{T}^d$ ($d\geq 2$).
%Let $g$ be the generation number in approximating $\nu_{c,\Delta t}$. 
%Each generation moves and replicates $m$-times, with a life span $T$ (time period), time step 
%$\Delta t = T/m$.   
%\While{$g=1:G-1$}{  
%\While{$j=0:m-1$}{
%$\boldsymbol{\eta}^{j}_{g} 
%\leftarrow$ one-step-advection-diffusion update on %$\boldsymbol{\xi}^{j}_{g}$.
%Define fitness $\boldsymbol{F}:= \exp\{ c(T- j\Delta t, %\boldsymbol{\eta}^{j}_{g})\, \Delta t \}$. 
%$E_{g,j}:= {1\over \Delta t} \ln$ (mean population fitness).
%Normalize fitness to weight $\boldsymbol{p}:= %\boldsymbol{F}/SUM(\boldsymbol{F})$.
%$\boldsymbol{\xi}^{j+1}_{g} \leftarrow $ resample 
%$\boldsymbol{\eta}^{j}_{g}$ via multinomial distribution with weight 
%$\boldsymbol{p}$.}
%$\boldsymbol{\xi}^{0}_{g+1} \leftarrow \boldsymbol{\xi}^{m}_{g}$, %$E_{g} \leftarrow  {\rm mean}\,( E_{g,j})$ over $j$.
%}
% Output: approximate $\mu_{\Delta t}\leftarrow $ mean($E_{g}$), and 
%$\boldsymbol{\xi}^{0}_{G}$.   
%\caption{Genetic Interacting Particle Method}\label{gIPM1}
%\end{algorithm}
\bigskip

\begin{figure}[htbp]
	\centering
	\subfigure[Ground truth invariant measure]{\includegraphics[width=0.44\linewidth]{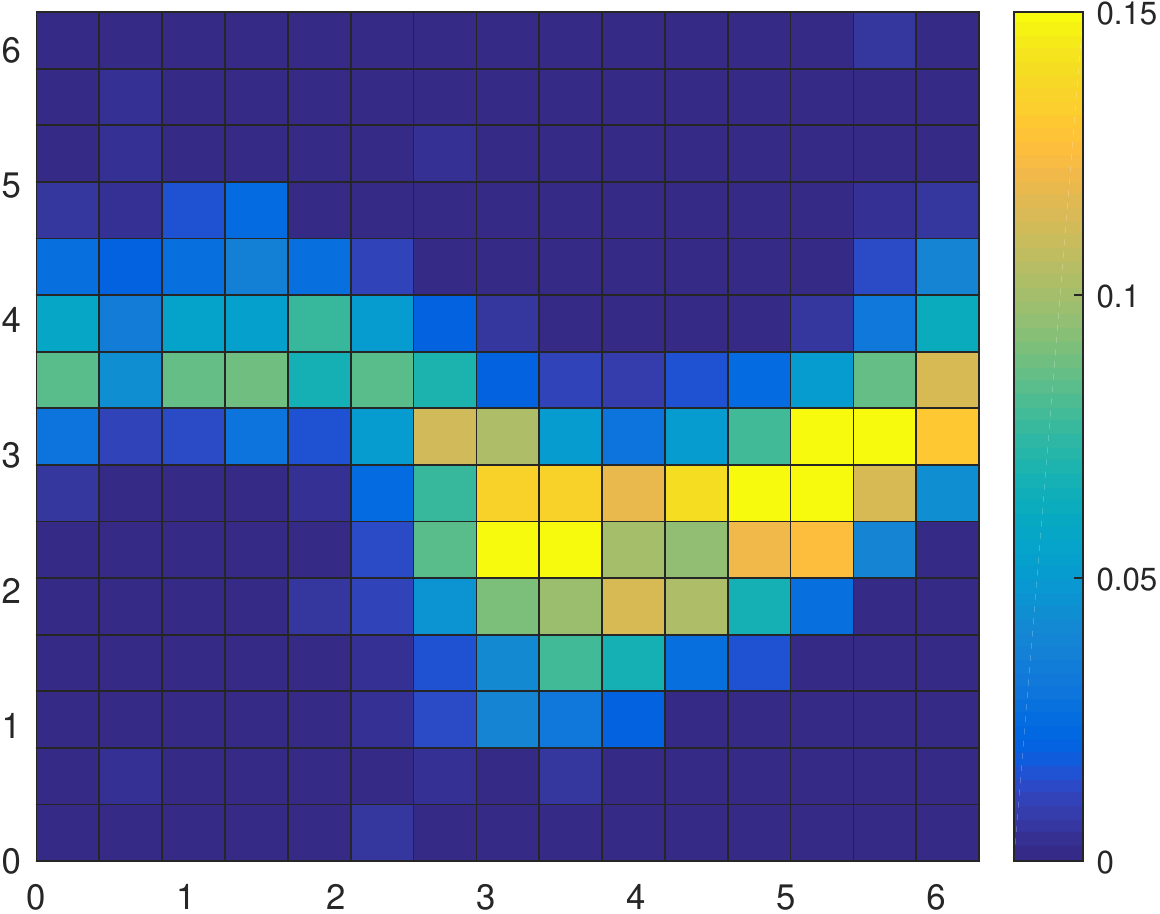}\label{fig:eg2d_ref}}
	\subfigure[Predicted invariant measure]{\includegraphics[width=0.44\linewidth]{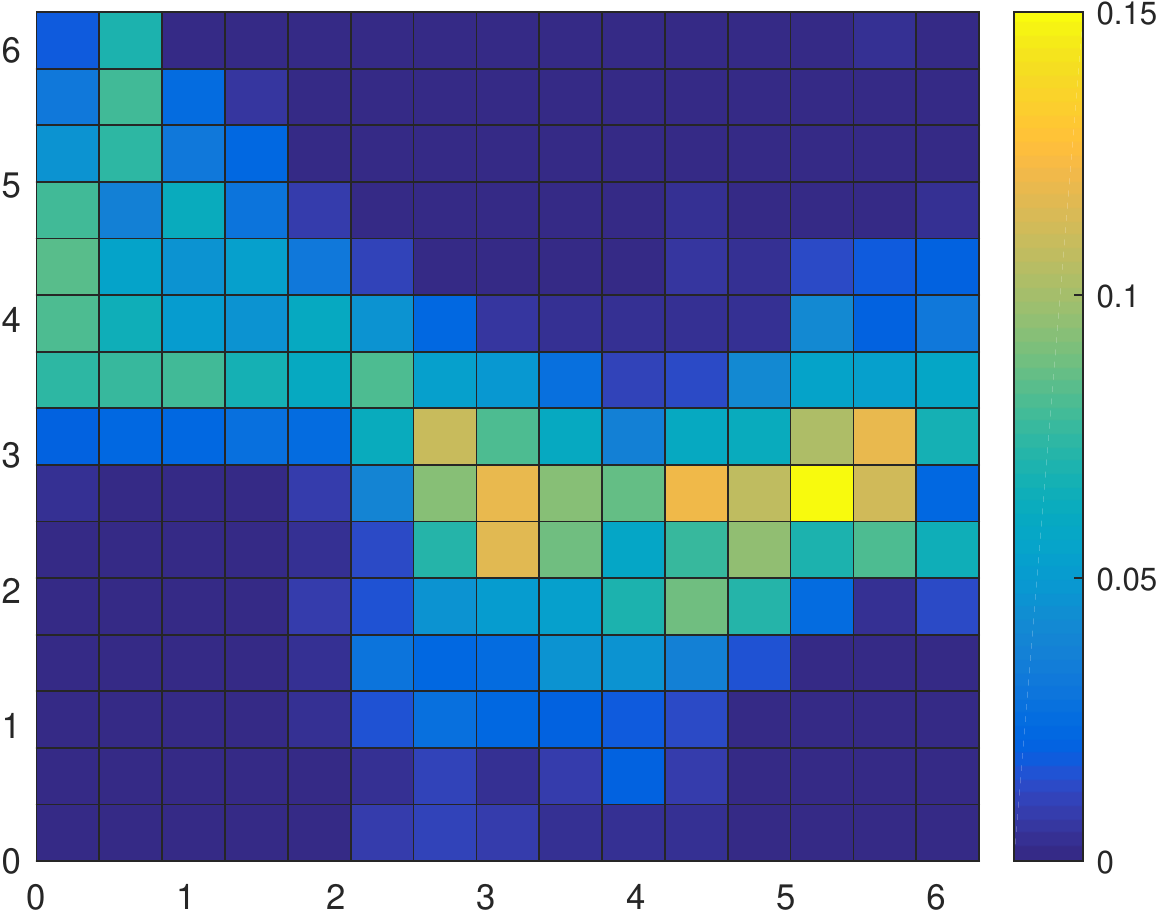}\label{fig:eg2d_sample_gen}}\\
	\subfigure[$W^2$ training error vs. gradient descent steps: spikes of height $\approx 0.1$ occur due to transition matrix $\gamma$ re-optimized in response to mini-batching of input training data.]{\includegraphics[width=0.44\linewidth]{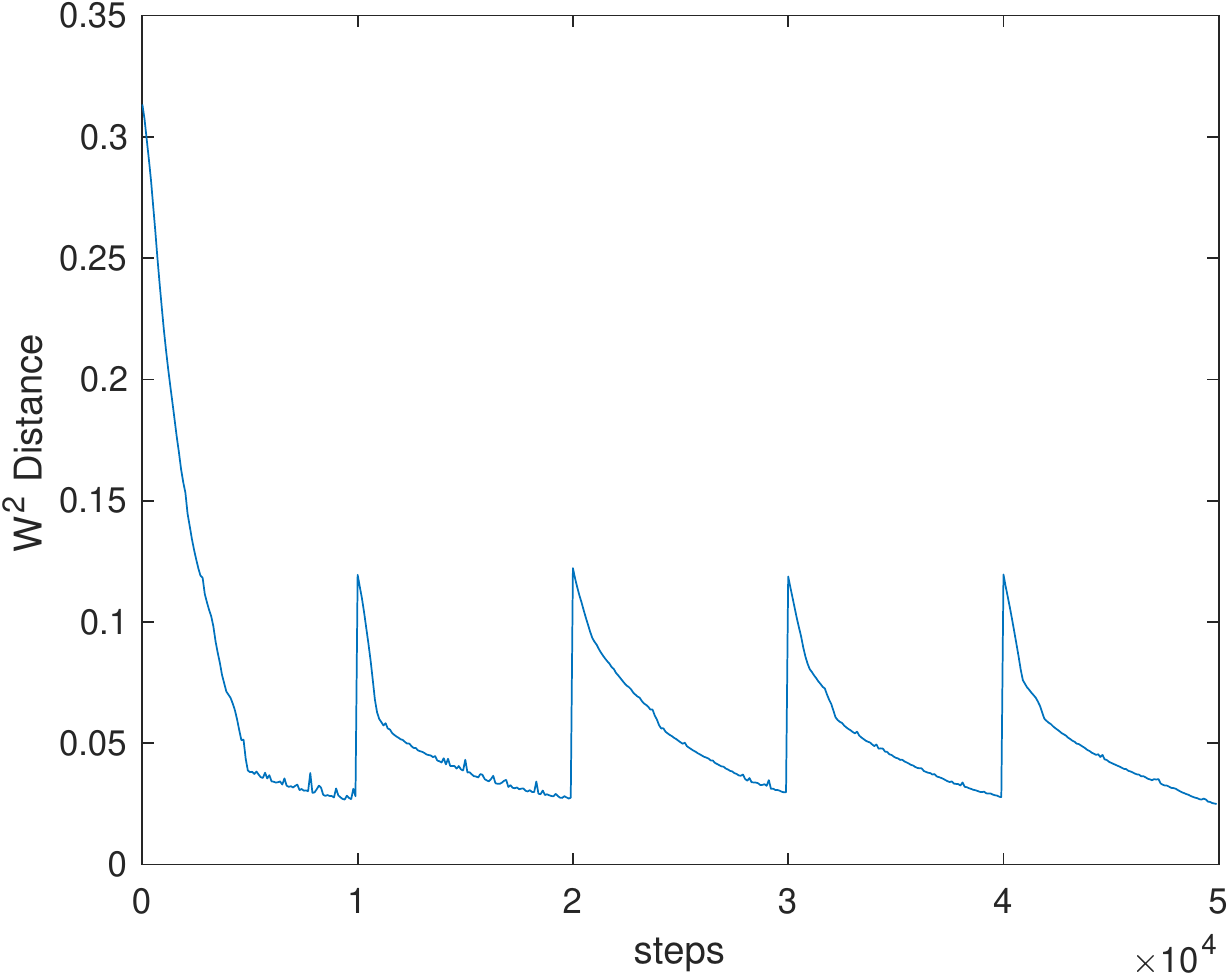}\label{fig:eg2d_w2_steps}}
	\subfigure[Acceleration: convergence to $\lambda_{\Delta t}$ value computed by Alg.\ref{gIPM1} with warm/cold start by DeepParticle prediction (red)/uniform distribution (blue).]{\includegraphics[width=0.44\linewidth]{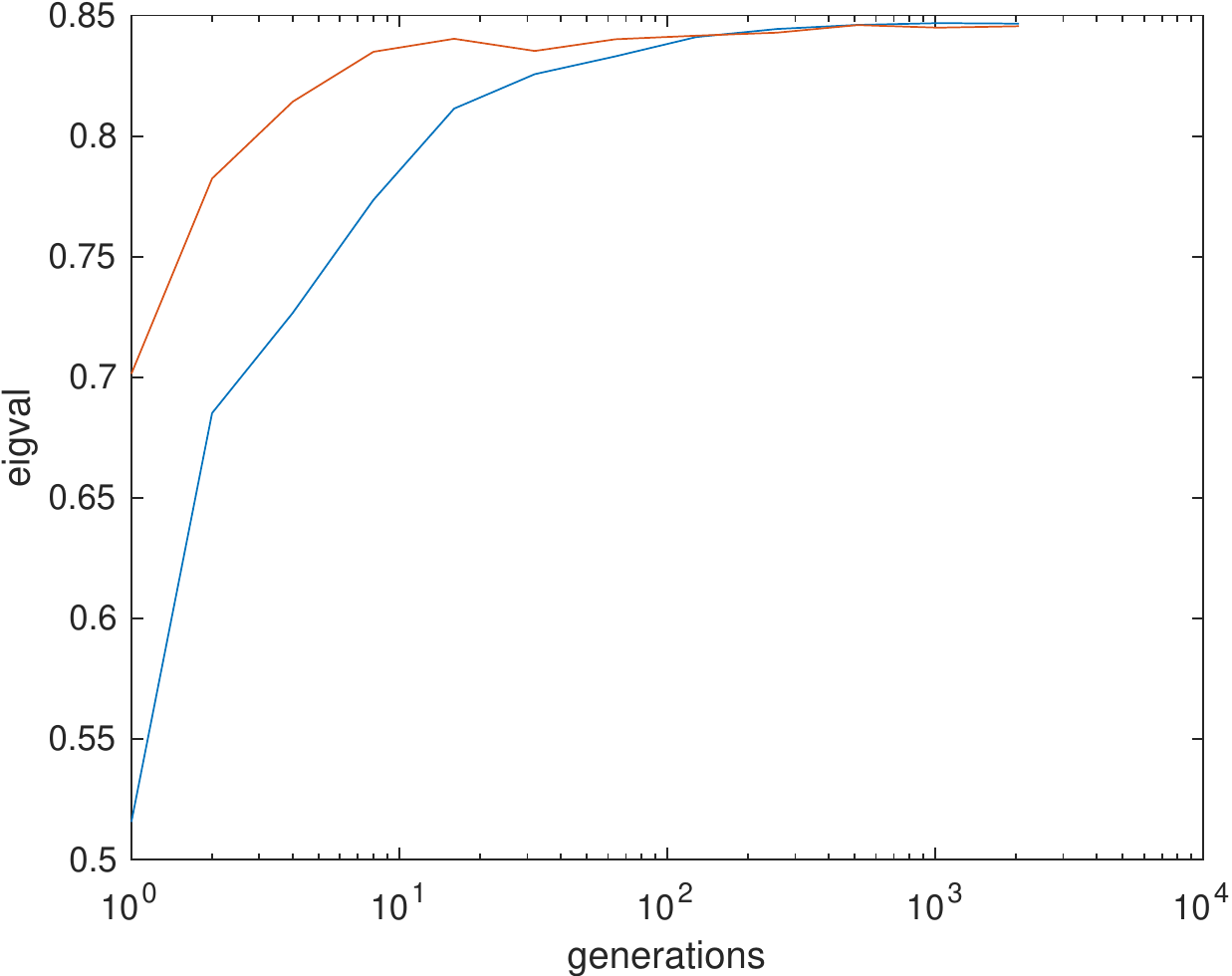}\label{fig:eg2d_eigcompare}}
	\caption{DeepParticle prediction (top), training/acceleration (bottom) at test value  $\kappa=2^{-4}$ in a 2D front speed computation. Loss reduction in c) shows fast (steps before 1.e4) and slow phases \cite{ZYX_21}.}
\label{fig:eg2d_prediction}
\end{figure}

%If the velocity field $\textbf{v}=\textbf{v}(\textbf{x})$ in  the KPP equation \eqref{KPP-eq} is time-independent, the minimal front speed in direction $\textbf{e}$ is given by the variational formula \cite{gartner1979}: $c^{*}(\textbf{e})=\inf_{\lambda>0}\mu(\lambda)/\lambda$, where $\mu(\lambda)$ is the principal eigenvalue of the elliptic operator,
%\begin{align}
% &\mathcal{L}\Phi=	\sigma \Delta_{\mathbf{x}} \Phi+(2 \sigma \lambda \mathbf{e}+\mathbf{v}) \cdot \nabla_{\mathbf{x}} \Phi+\left(\sigma \lambda^{2}+\lambda \mathbf{e} \cdot \mathbf{v}+\tau^{-1} f^{\prime}(0)\right) \Phi=\mu(\lambda) \Phi, \label{elliptic-op} \\
% &\mathbf{x} \in D=[0,2 \pi]^{2}, %\label{elliptic-op-bc}
% \end{align}
%In Eq.\eqref{elliptic-op}, $\Phi \in L^2(D)$ and $\textbf{v}$ is period $1$ in each coordinate direction $x_i$, $1 \le i \le d$.
%Accurate estimation of $c^{*}(\textbf{e})$ boils down to computing the principal eigenvalue of the operator $\mathcal{A}^{\lambda}_{1}$ in \eqref{elliptic-op}. Adaptive finite element methods (FEM) were successfully applied to solve \eqref{elliptic-op} in \cite{shen2013finite2d,shen2013finite3d}. 
{ An advantage of the IPM is that  given the same particle number, the computational cost of generating training data scales linearly with the dimension of spatial variables. As a comparison, in the Eulerian framework, one needs to solve the principal eigenvalue of a parabolic operator $\partial_t-\mathcal{A}$ by discretizing the eigenvalue problem on mesh grids. The number of mesh grids depends exponentially on the spatial dimension, which becomes expensive for 3D problems.
Though adaptive mesh techniques can alleviate the computational burden, their  design and implementation are challenging for time dependent 3D problems when large gradient regions are dynamically changing and repeated mesh adaptations must be performed. In contrast, the IPM is spatially mesh-free and self-adaptive.} 
The computational bottleneck remains in IPM at small $\kappa$ when we need a large number of particles running for many generations to approach the invariant measure. To address this issue, we apply our DeepParticle method to generate particle samples by initiating at {\it a learned distribution} resembling the invariant measure and accelerating Alg. \ref{gIPM1} (i.e. reduce $G$ to reach convergence). 
\medskip
\paragraph{2D steady cellular flow}
We first consider a 2D cellular flow   $\mathbf{v}=\left(-\sin x_{1} \cos x_{2}, \cos x_{1} \sin x_{2}\right)$. In this case, we apply the physical parameter dependent network described in Section \ref{subsec:parNN} to learn the invariant measure corresponding to different $\kappa$ values. To generate training data, we first use the IPM, Alg. \ref{gIPM1} with $\kappa_i=2^{-2-0.25*(i-1)}$, $i=1: 8 (= n_{\eta})$, and $N_0=40000$ particle evolution for $G=2048$, $\Delta t=2^{-8}$, $T=1$,  to get samples of invariant measure at different $\kappa$ values. The value of $G$ is chosen so that the direct IPM simulation of principal eigenvalue converges, see blue line of Fig.\ref{fig:eg2d_eigcompare}. From each set of sample points with different $\kappa$'s, we randomly pick $N=2000$ sample points without replacement, denoted as $\mathcal{Y}_1, \cdots,\mathcal{Y}_8$.  Under such setting, we then seek neural networks $f_\theta$ such that given $\kappa_i$ and a set 
$\mathcal{X}_i$ of i.i.d. uniformly distributed points on $[0,2\pi]^2$, the network output $f_\theta(\mathcal{X}_i;\kappa_i)$ is distributed near $\mathcal{Y}_i$ in $W^2$ distance. 
%Here $\mathcal{X}_i$ can be selected from process of generating $\mathcal{Y}_i$ or randomly generate from a known distribution.
The set $\{\mathcal{Y}_1, \cdots,\mathcal{Y}_8\}$ is called one mini-batch of training data and in total we have $N_{dict}=5$ mini-batches through $50000$ steps of network training.  As stated in Alg. \ref{alg:1}, after each mini-batch of network training (10000 steps of gradient descent), we re-optimize $\gamma$ until its normalized Frobenius norm is greater than $tol=0.7$. Note that only a quarter of the 40,000 particles in the IPM 
simulations have been used in training.
\medskip

During training, we apply Adams gradient descent with initial learning rate $0.002$ and a weight decay %an $L_2$ regularizer 
with hyper-parameter $0.005$ for trainable weights in the network. In each step, we select $M=25$ columns and rows by Alg. \ref{alg:pick_col} in solving the sub-optimization problem \eqref{def:lp-sub-cost} and repeat the linear programming for $N_{LP}=10$ times.
\medskip

\begin{figure}[htbp]
	\centering
	\subfigure[$\kappa=2^{-2}$]{\includegraphics[width=0.24\linewidth]{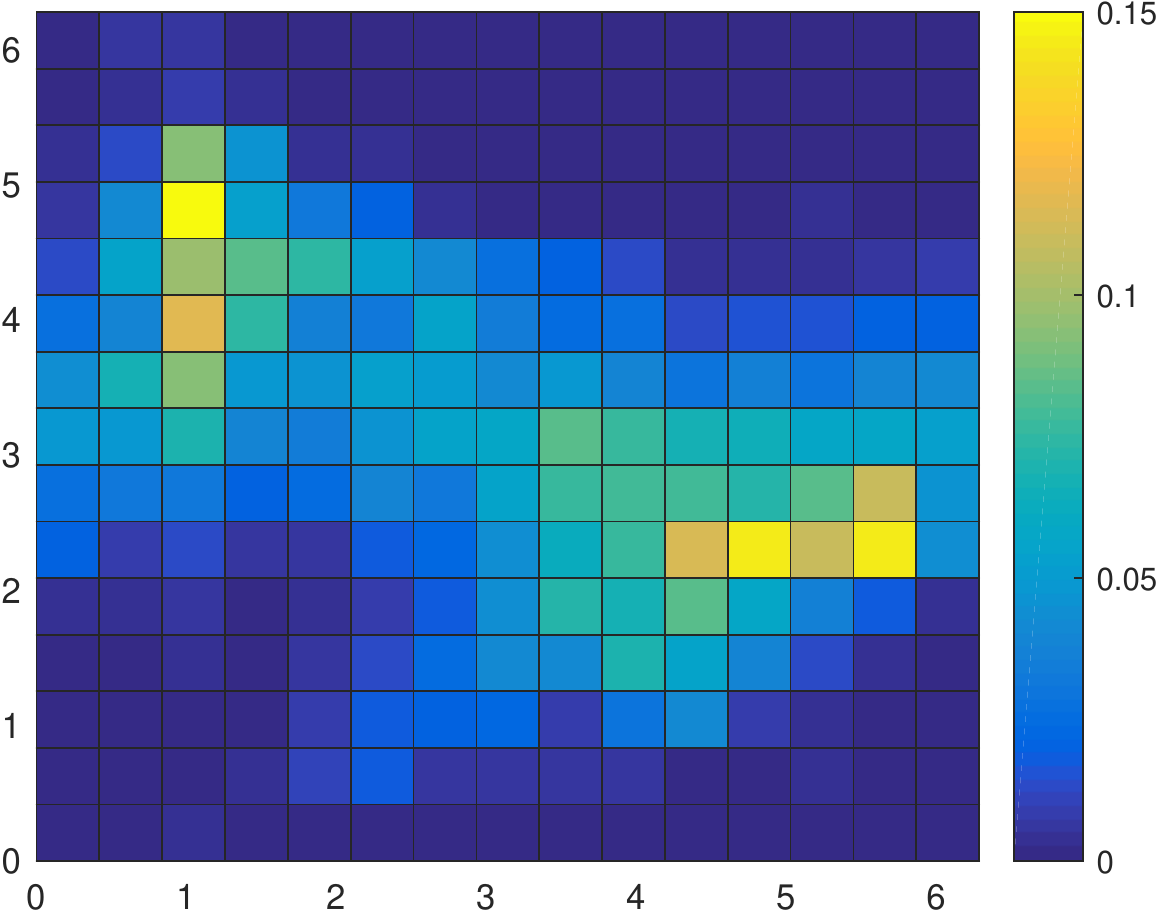}}
		\subfigure[$\kappa=2^{-2.5}$]{\includegraphics[width=0.24\linewidth]{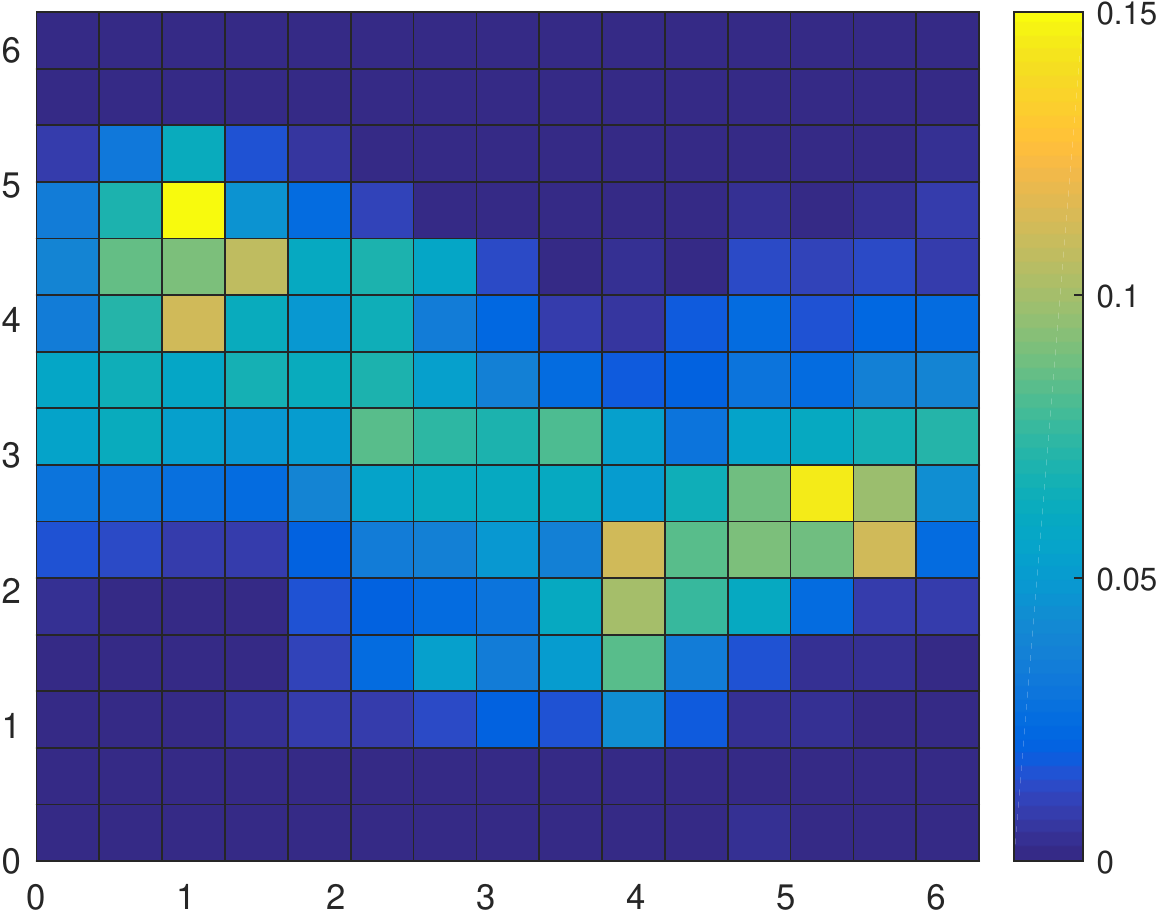}}
	\subfigure[$\kappa=2^{-3}$]{\includegraphics[width=0.24\linewidth]{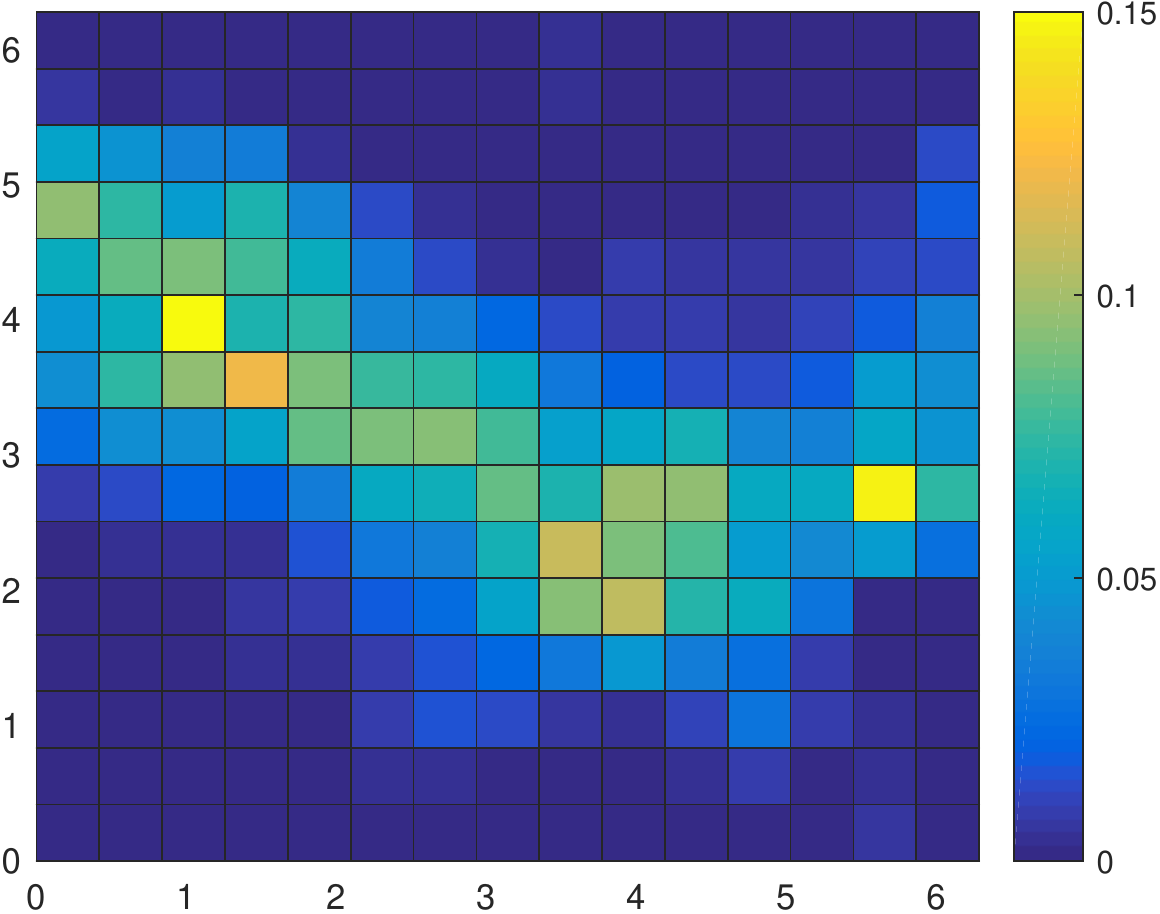}}
		\subfigure[$\kappa=2^{-3.5}$]{\includegraphics[width=0.24\linewidth]{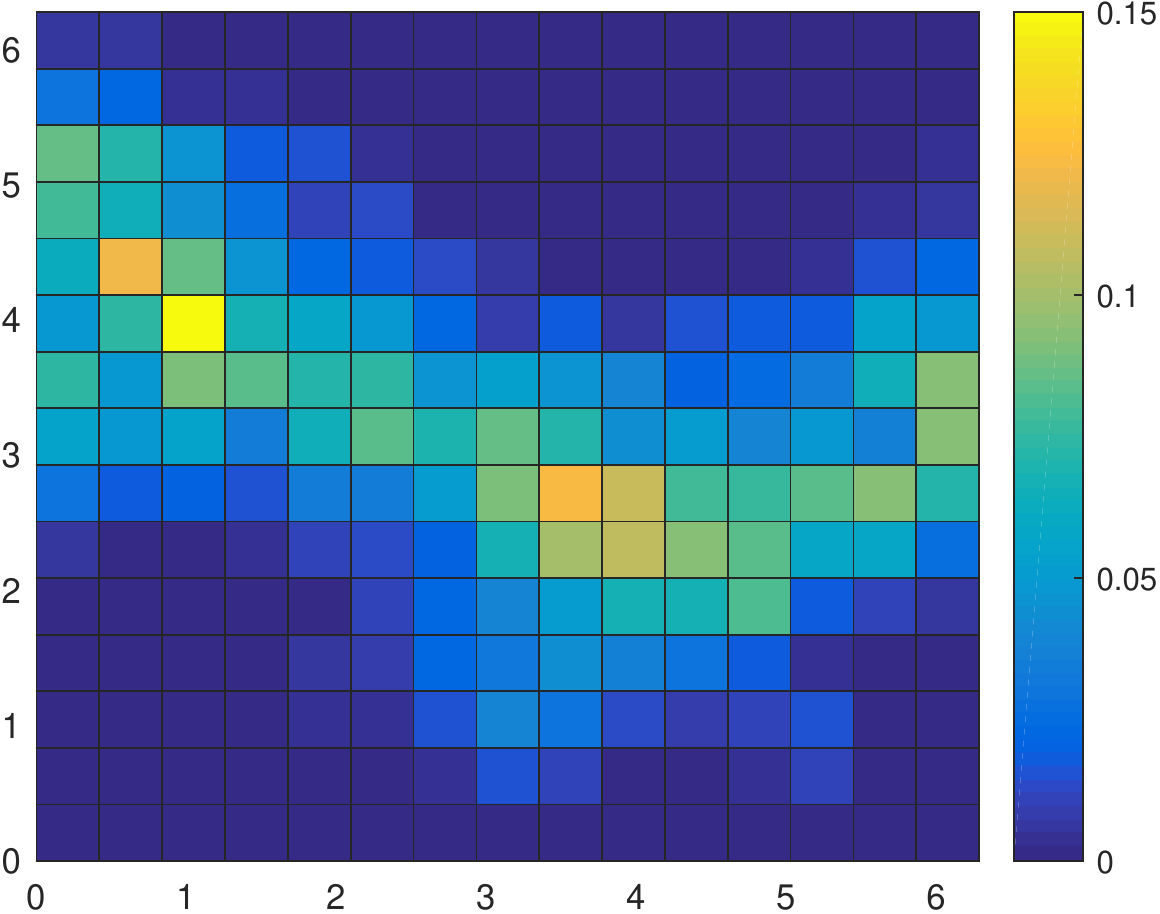}}
\\
	\subfigure[$\kappa=2^{-2}$]{\includegraphics[width=0.24\linewidth]{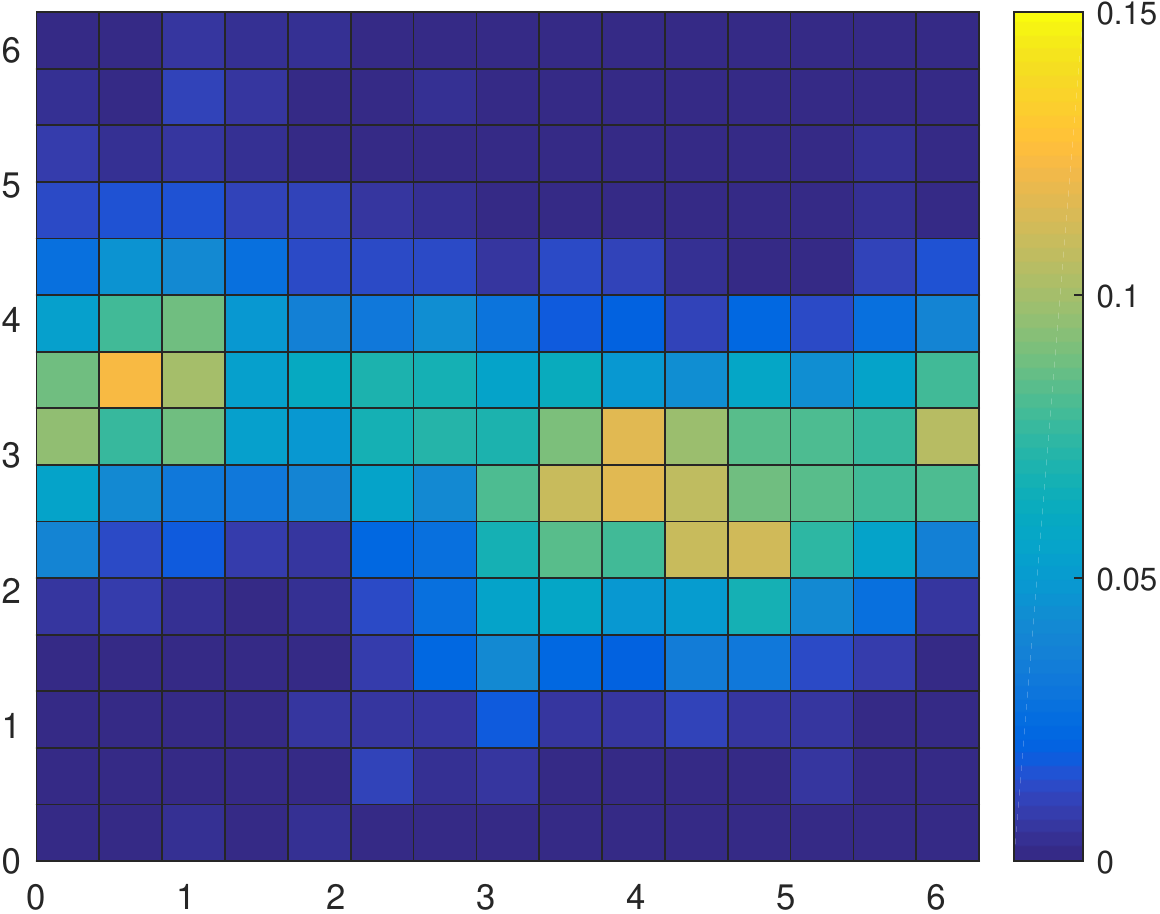}}
	\subfigure[$\kappa=2^{-2.5}$]{\includegraphics[width=0.24\linewidth]{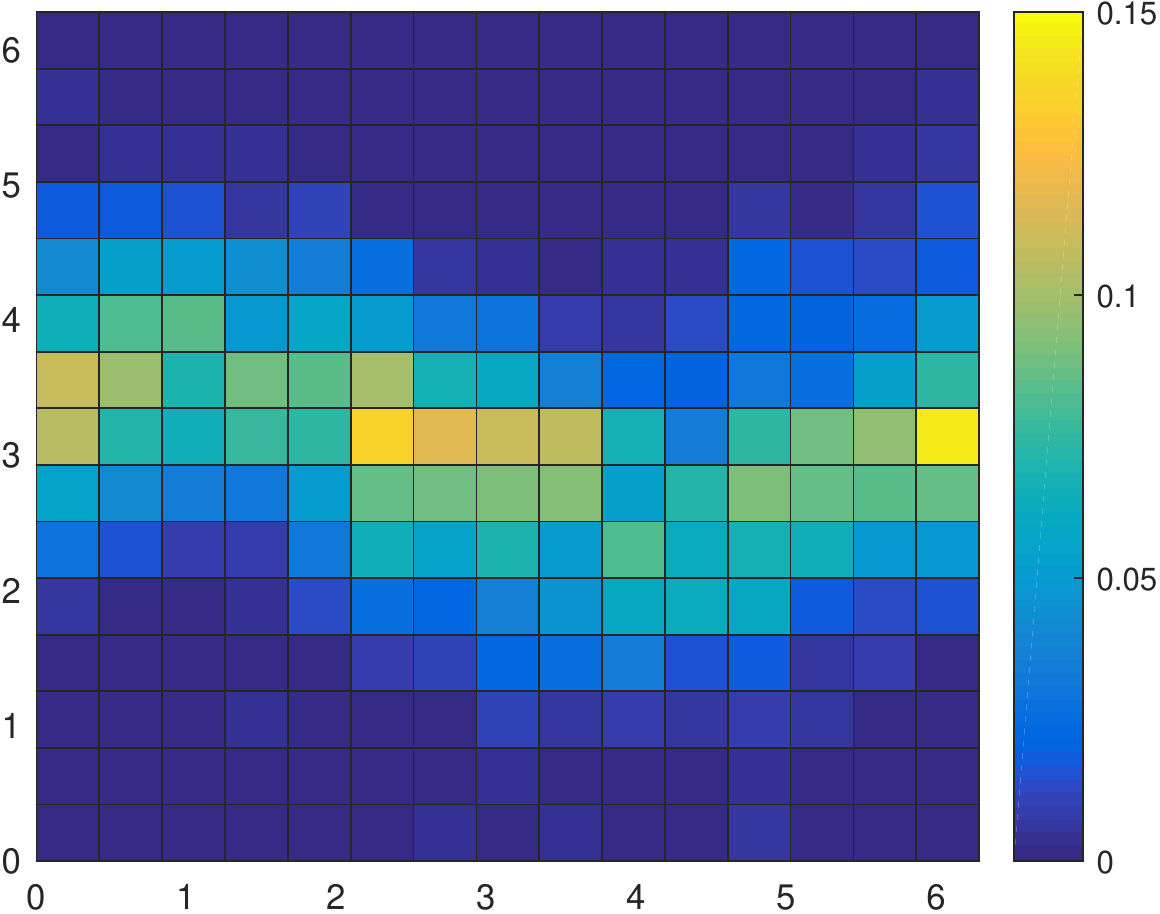}}
	\subfigure[$\kappa=2^{-3}$]{\includegraphics[width=0.24\linewidth]{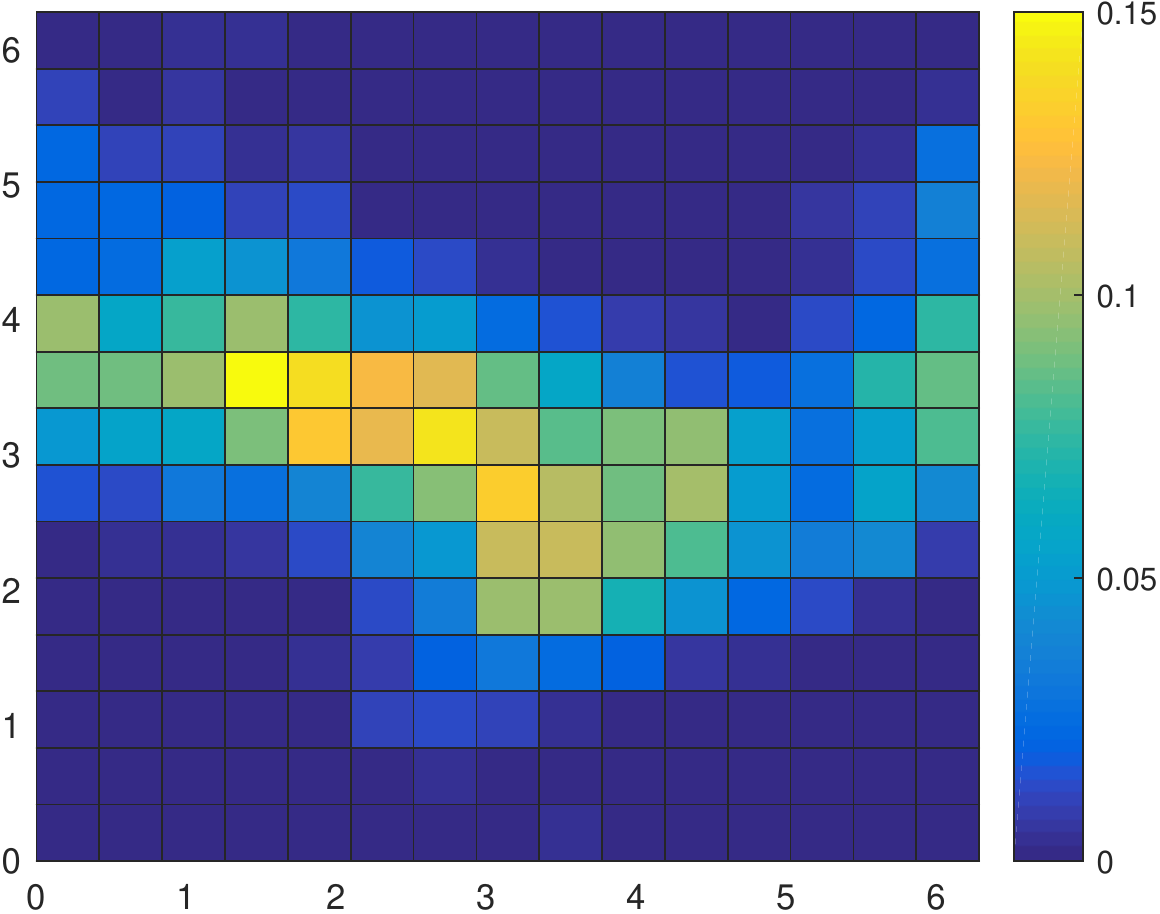}}
	\subfigure[$\kappa=2^{-3.5}$]{\includegraphics[width=0.24\linewidth]{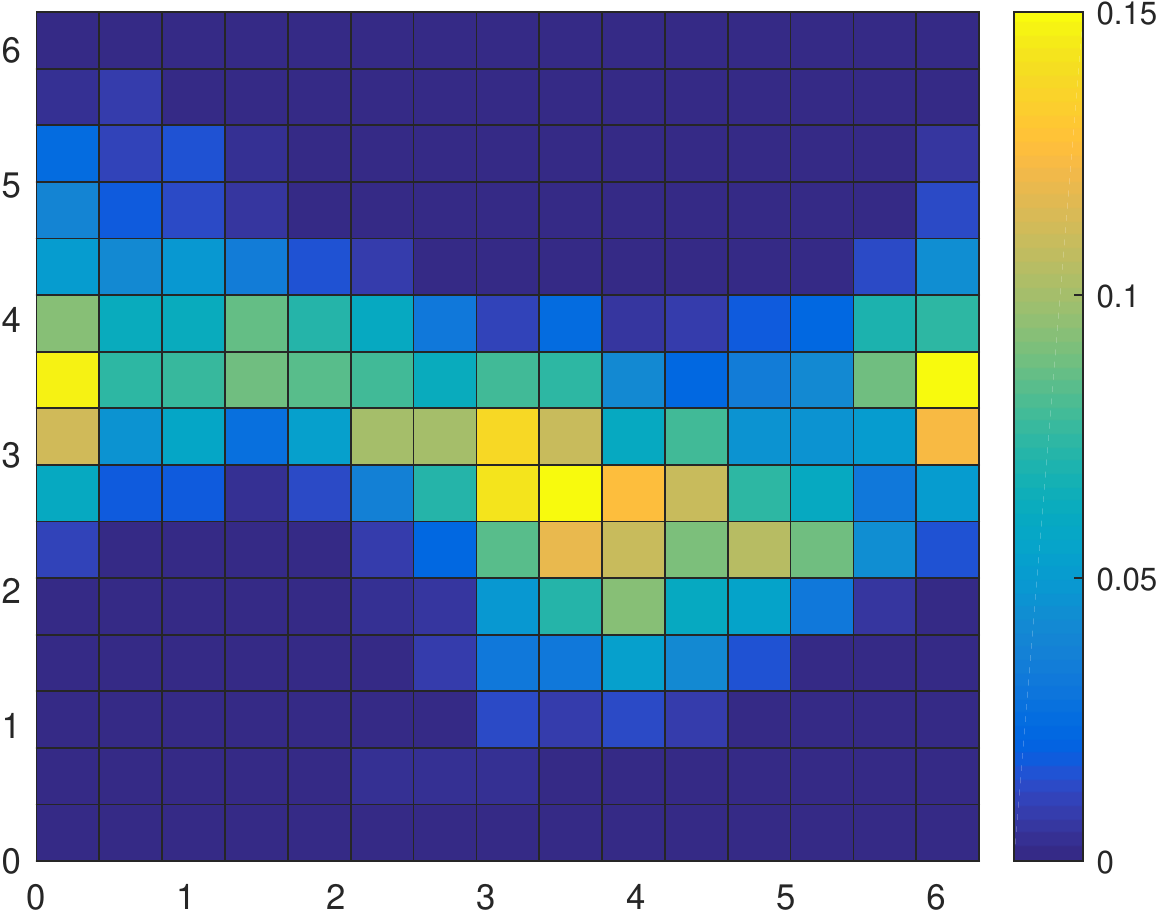}}
	
	\caption{Generated invariant measure (upper row) and corresponding training data (reference, lower row)}
	\label{fig:eg2d_kappa}
\end{figure}

In Fig. \ref{fig:eg2d_kappa}, we present the performance of our algorithm in sampling distribution in the original training data. 
 Each time, the training algorithm only has access to at most $N\times N_{dict}=10000$ samples of each target distribution with $\kappa_i=2^{-2-0.25*(i-1)}$, $i=1, \cdots, 8= n_{\eta}$. After training, we generate $N_0=40000$ uniformly distributed points as input of the network. We compare the histogram of network output  and that of the $N_0=40000$ sample points obeying the invariant measure from the IPM Alg.\ref{gIPM1}. We see that our trained parameter-dependent network indeed reproduces the sharpening effect on the invariant measure when $\kappa$ becomes small.

In Fig. \ref{fig:eg2d_prediction}, we present the performance of algorithm in predicting the invariant measure of IPM with $\kappa=2^{-4}$.  First we compare the invariant measure generated from direct simulation with IPM (Fig. \ref{fig:eg2d_ref}) and from the trained network (Fig. \ref{fig:eg2d_sample_gen}). In Fig. \ref{fig:eg2d_w2_steps}, we show the $W_2$ distance between the generated distribution and the target distribution of the training data. The $W_2$ distance goes up when we re-sample the training data. This is because at this moment, the transition matrix $\gamma$ in the  definition of the $W_2$ distance is inaccurate. Note that the test $\kappa=2^{-4}$ value is no longer within the training range $\kappa=2^{-p}$, $p\in [2,3.75]$. Our trained network predicts an invariant measure. Such prediction also serves as a `warm start' for IPM and can accelerate its computation to quickly reach a more accurate invariant measure. As  the invariant measure has no closed-form analytical solution, we plot in Fig. \ref{fig:eg2d_eigcompare} the principal eigenvalue approximations by IPM with warm and cold starts, i.e. one generated from DeepParticle network (warm) and the i.i.d. samples from a uniform distribution (cold). The warm start by DeepParticle achieves 4x to 8x speedup.

\begin{figure}
    \centering
    \subfigure[GD  steps = 1000]{\includegraphics[width=0.32\linewidth]{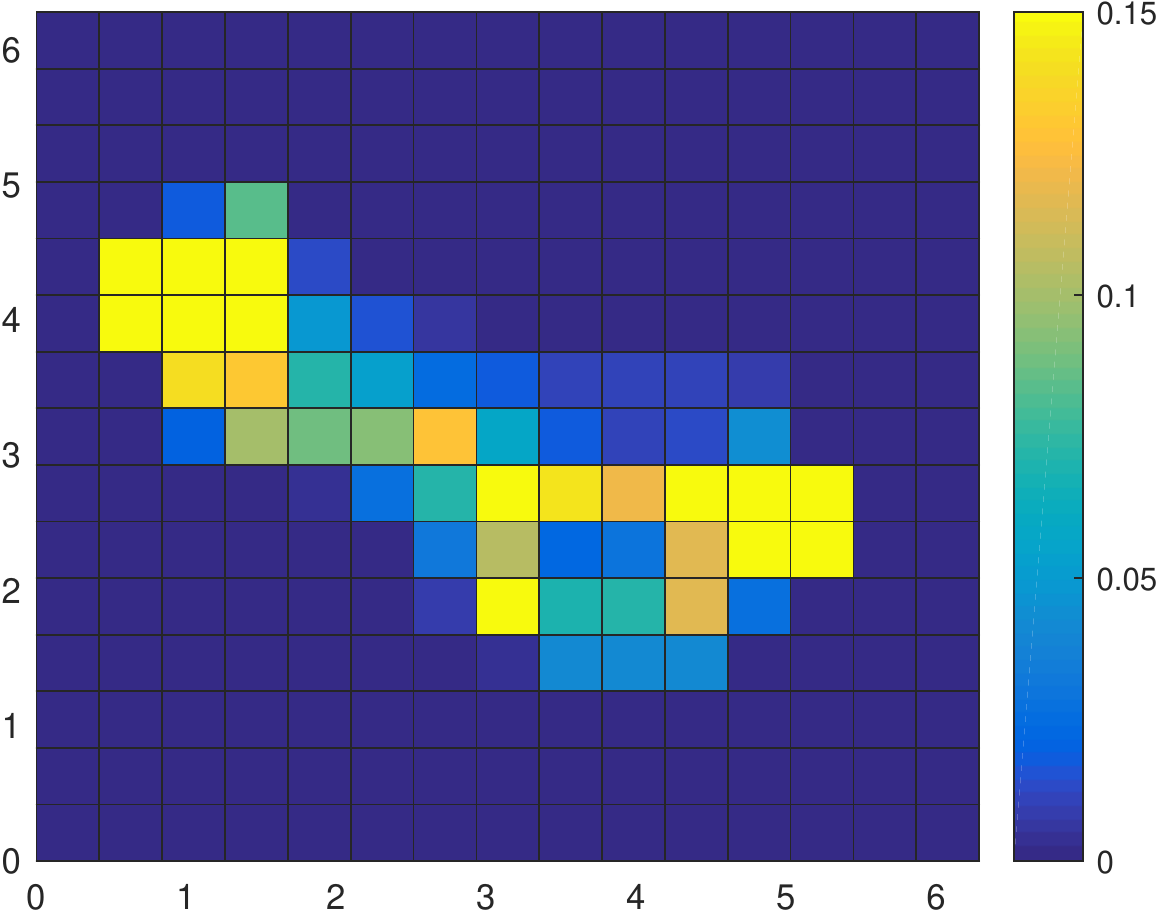}}
    \subfigure[GD steps = 4000]{\includegraphics[width=0.32\linewidth]{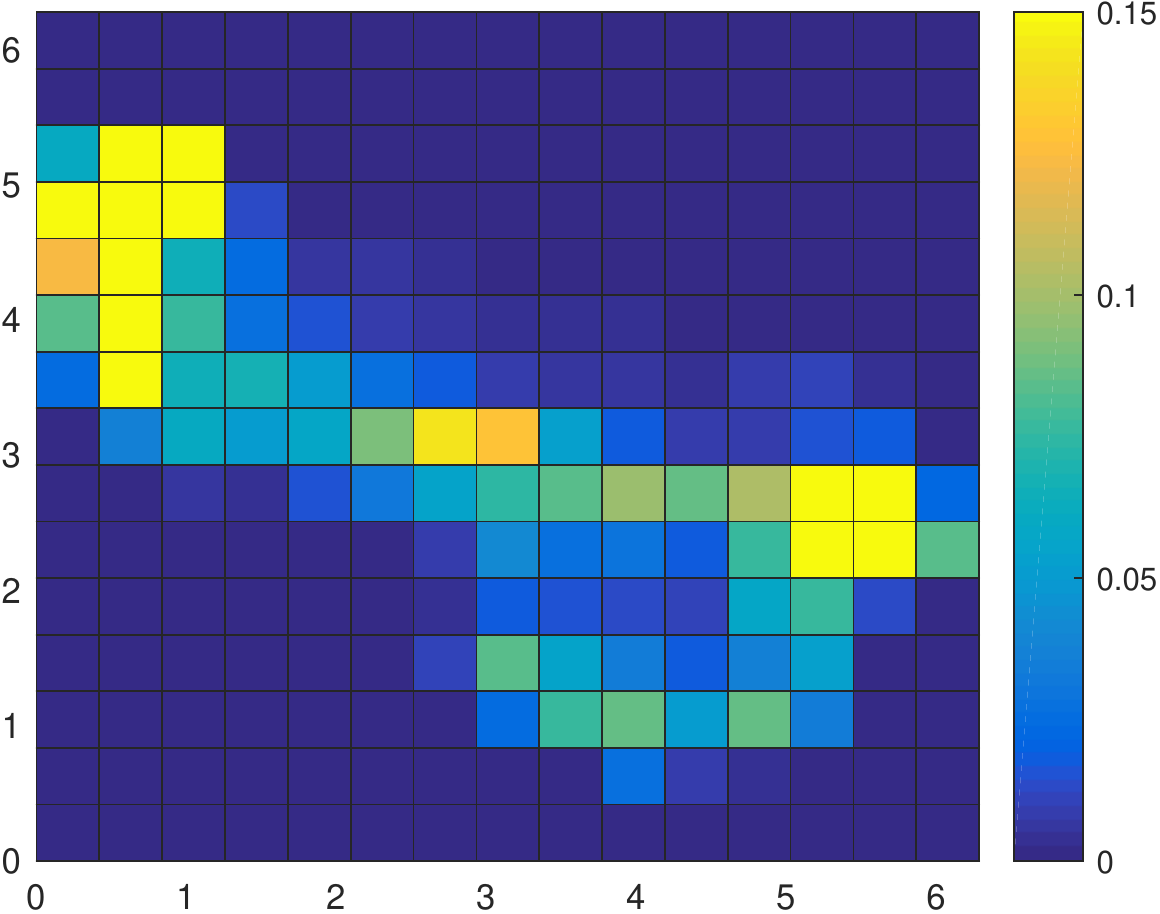}}
    \subfigure[GD steps  = 7000]{\includegraphics[width=0.32\linewidth]{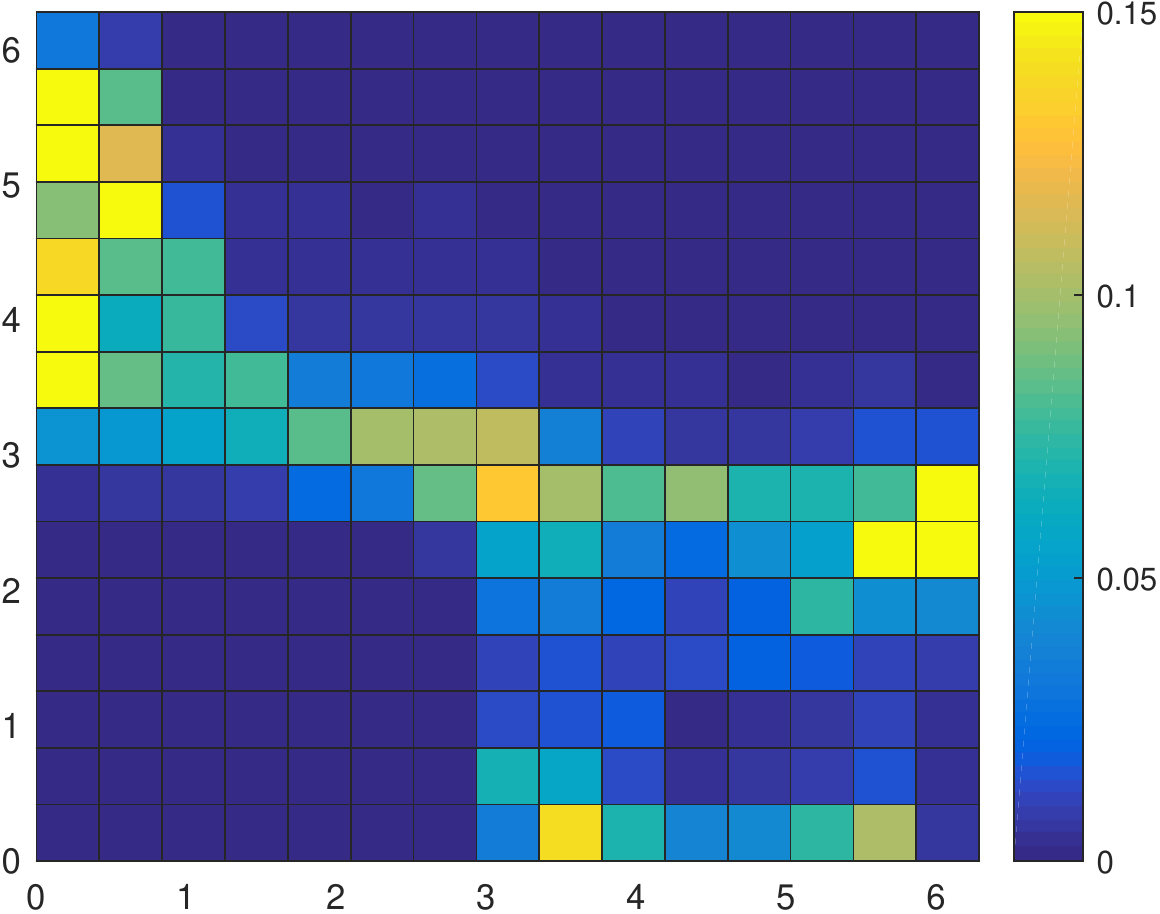}}
    \\
    \subfigure[GD steps = 10000]{\includegraphics[width=0.32\linewidth]{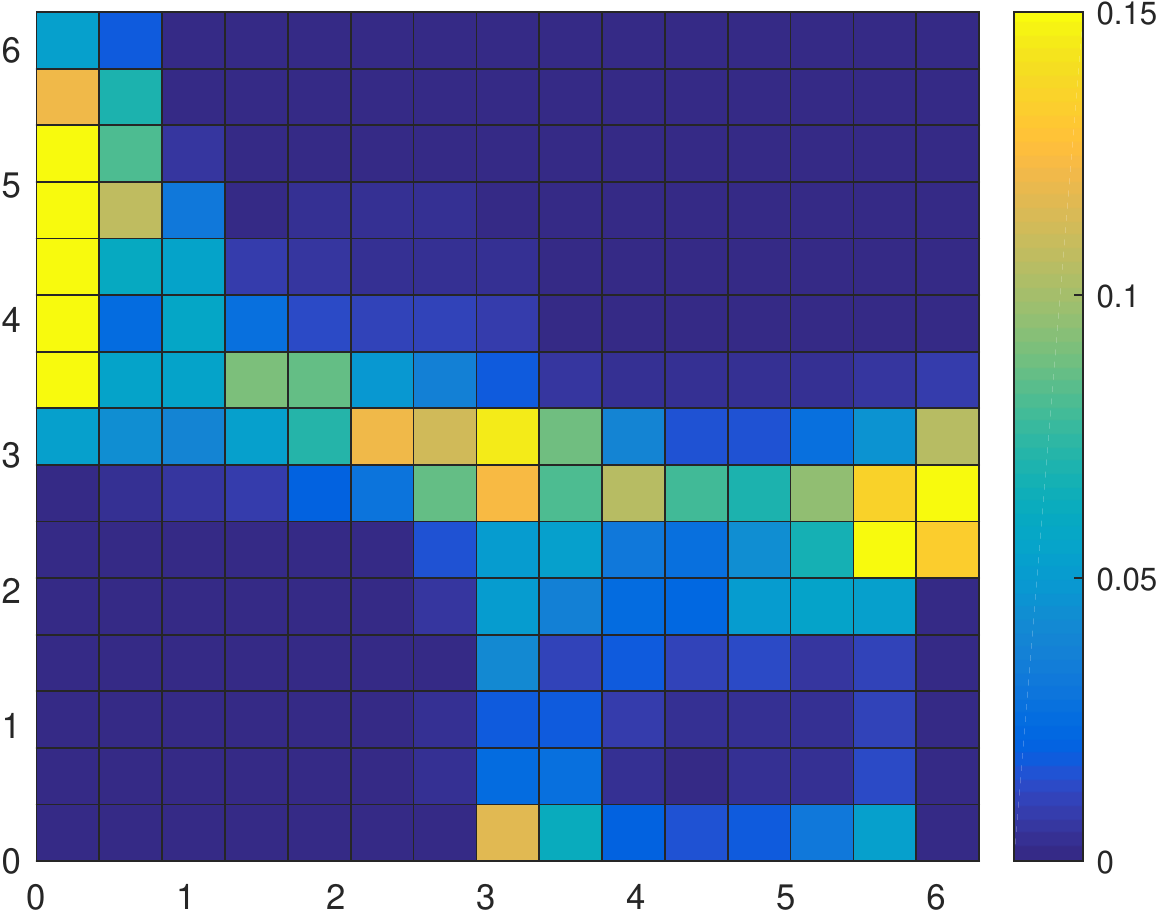}\label{fig:eg2d_log_10k}}
    \subfigure[GD steps = 15000]{\includegraphics[width=0.32\linewidth]{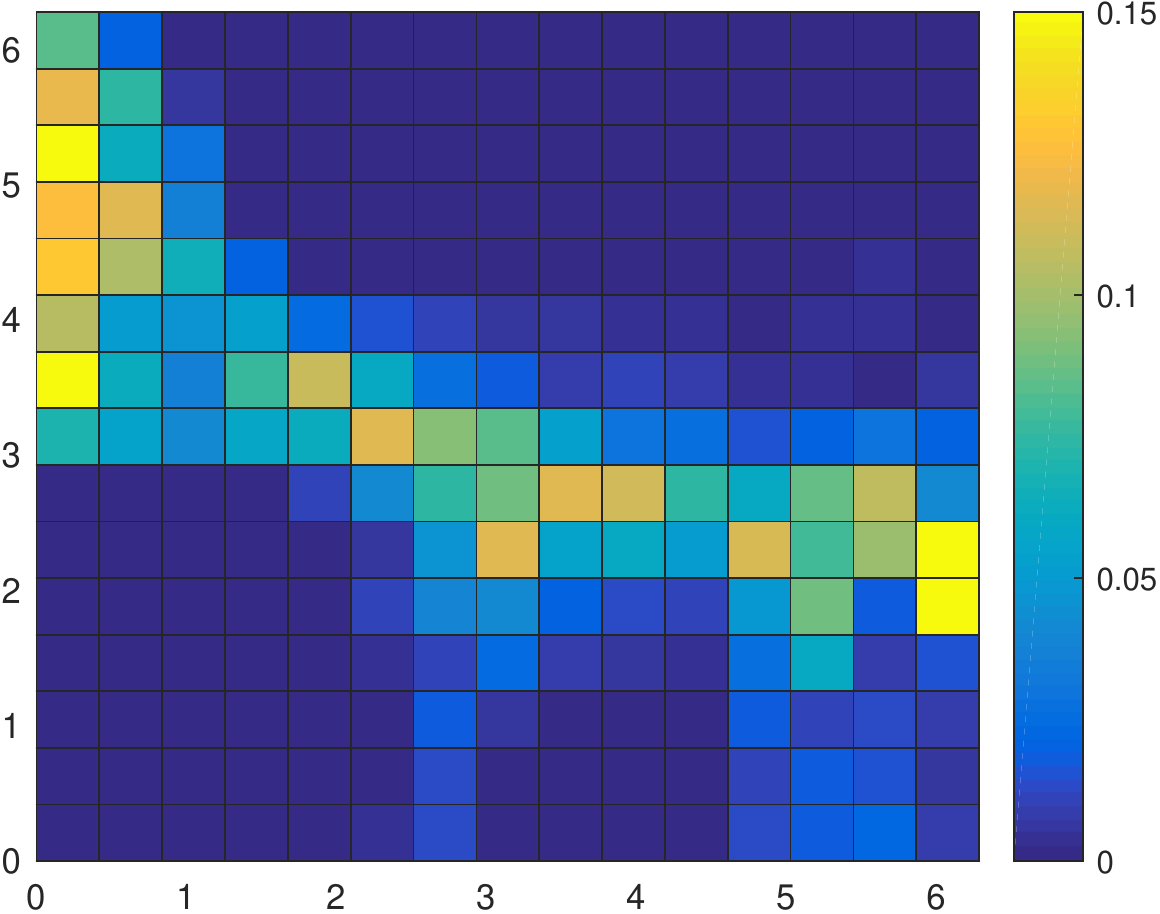}\label{fig:eg2d_log_15k}}
    \subfigure[GD steps = 20000]{\includegraphics[width=0.32\linewidth]{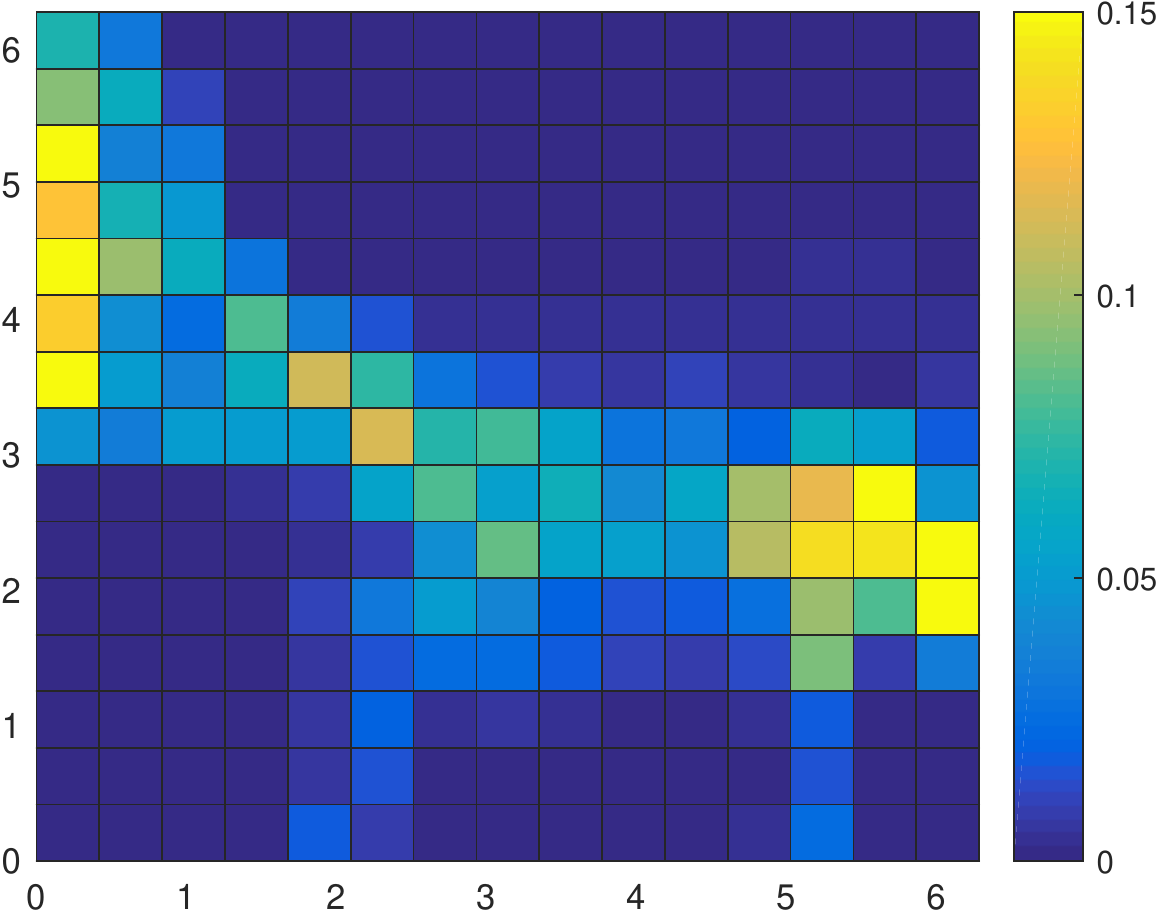}}
    \\
    \subfigure[GD steps = 
    30000]{\includegraphics[width=0.32\linewidth]{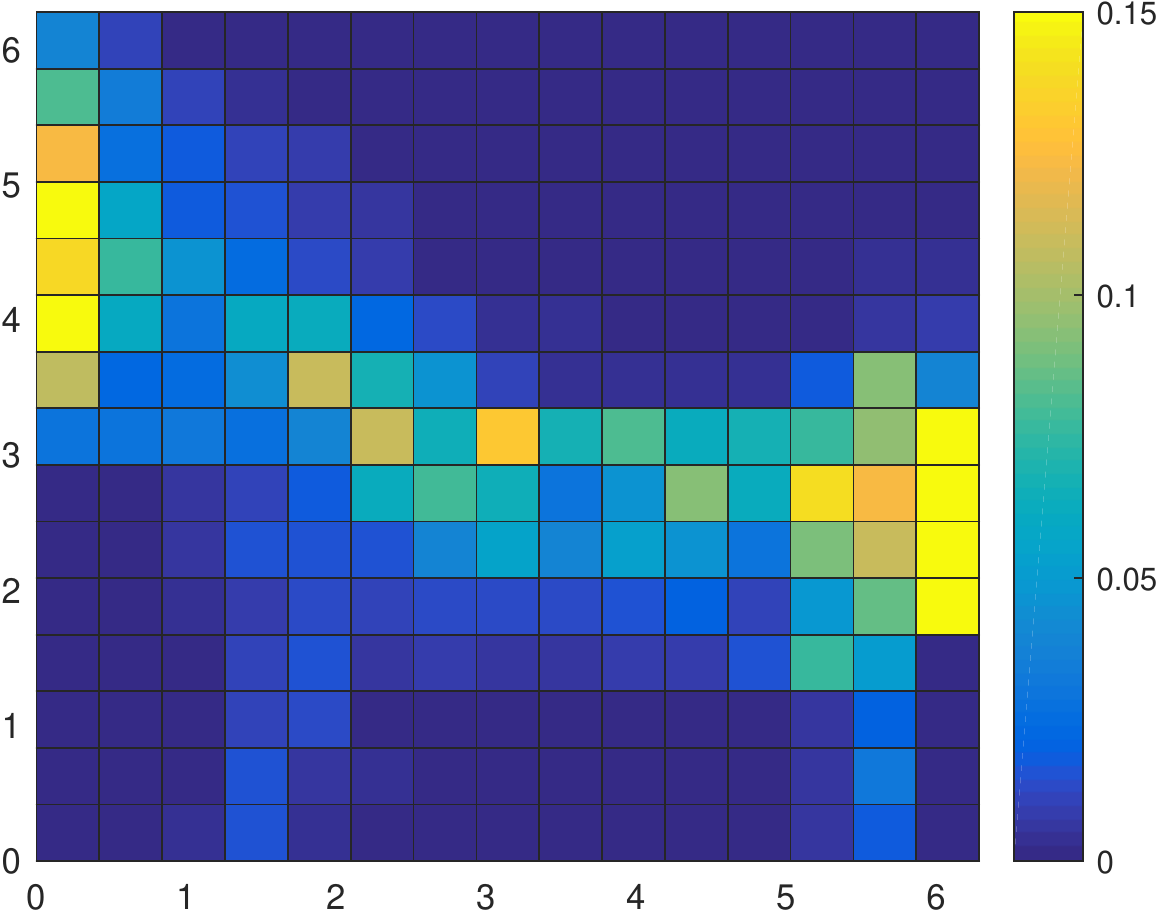}}
    \subfigure[GD steps = 40000]{\includegraphics[width=0.32\linewidth]{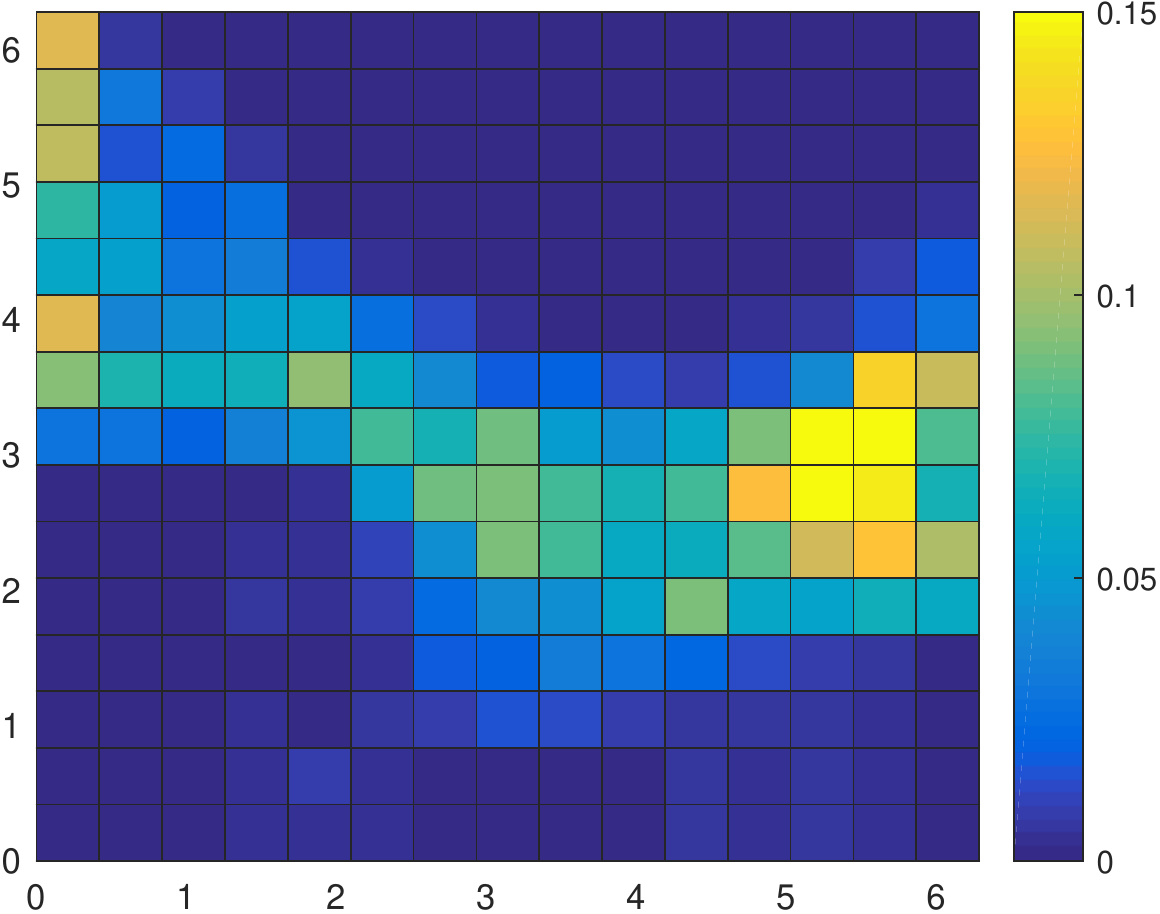}}
    \subfigure[GD  steps = 50000]{\includegraphics[width=0.32\linewidth]{fig/sigma_dependent/sample_gen-eps-converted-to.pdf}\label{fig:eg2d_log_50k}}
    \caption{Measures generated at different gradient descent (GD)  steps in network training for  KPP front speeds in 2D steady cellular flows.}
	\label{fig:eg2d_log}
\end{figure}

Fig. \ref{fig:eg2d_log} displays histograms of $N_0=40000$ samples generated from our DeepParticle algorithm at different stage of training. In Fig. \ref{fig:eg2d_log_10k} with only one set of data, we already get a rough prediction of target distribution. With more data sets added, the prediction quality improves; see Fig. \ref{fig:eg2d_log_15k} to Fig. \ref{fig:eg2d_log_50k}. 
\paragraph{3D time-dependent Kolmogorov flow}
Next, we compute KPP front speed in
a 3D time-dependent Kolmogorov flow \cite{galloway1992numerical}:
\[ \mathbf{v}= \left(\sin\left(x_3 +  \sin(2\pi t)\right), \sin\left(x_1 + \sin(2\pi t)\right), \sin\left(x_2 + \sin(2\pi t)\right)\right). \]
%where $\lambda=0.5$ and $\theta=1$. 
The hyper-parameter setting for network training remains the same as in the 2D case.  The physical parameter-dependent network in Section \ref{subsec:parNN} learns the invariant measure corresponding to eight different $\kappa$ training values. Besides the input $\mathbf{x}$ and output $\mathbf{f(x)}$ are both in 3 dimensions, the network layout and training procedure are the same as in the 2D example. 
\medskip

\begin{figure}[htbp]
	\centering
	\subfigure[$\kappa=2^{-2}$]{\includegraphics[width=0.24\linewidth]{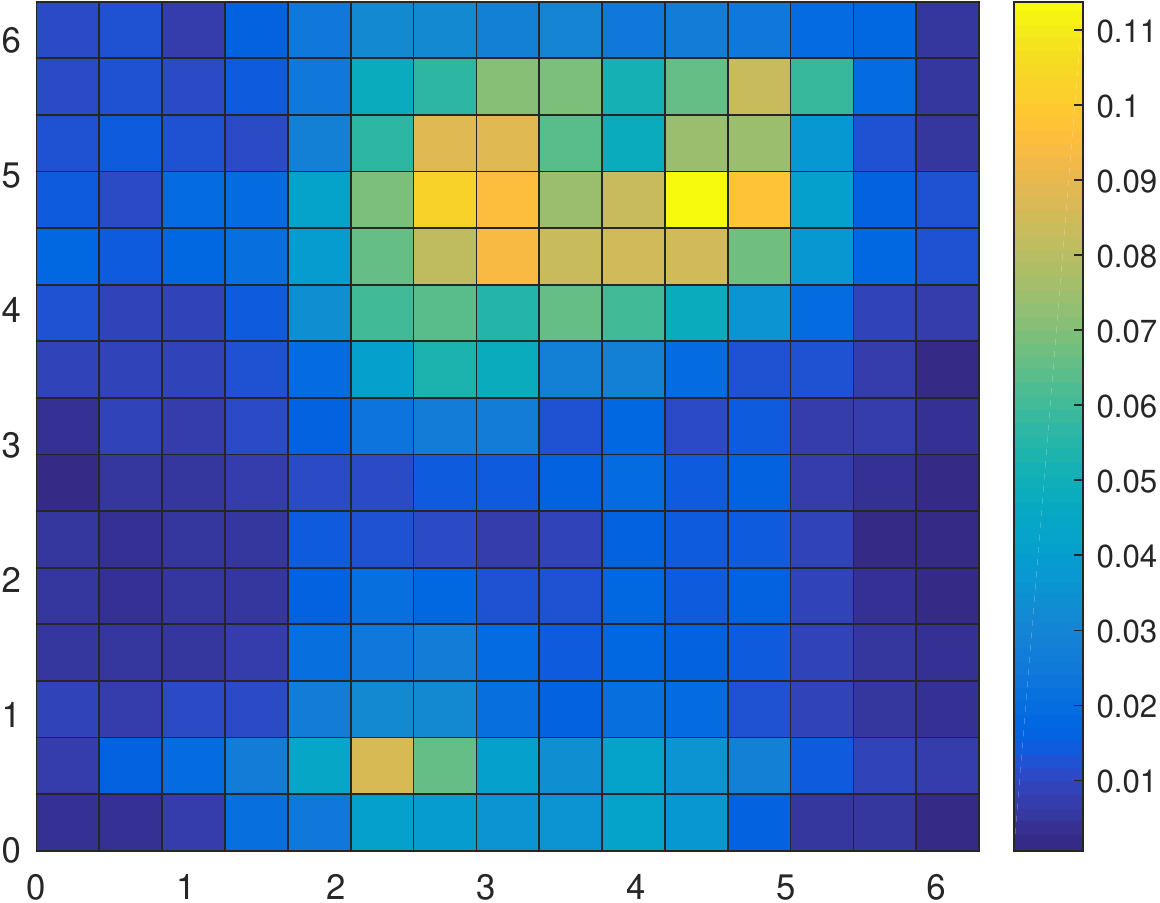}}
	\subfigure[$\kappa=2^{-2.5}$]{\includegraphics[width=0.24\linewidth]{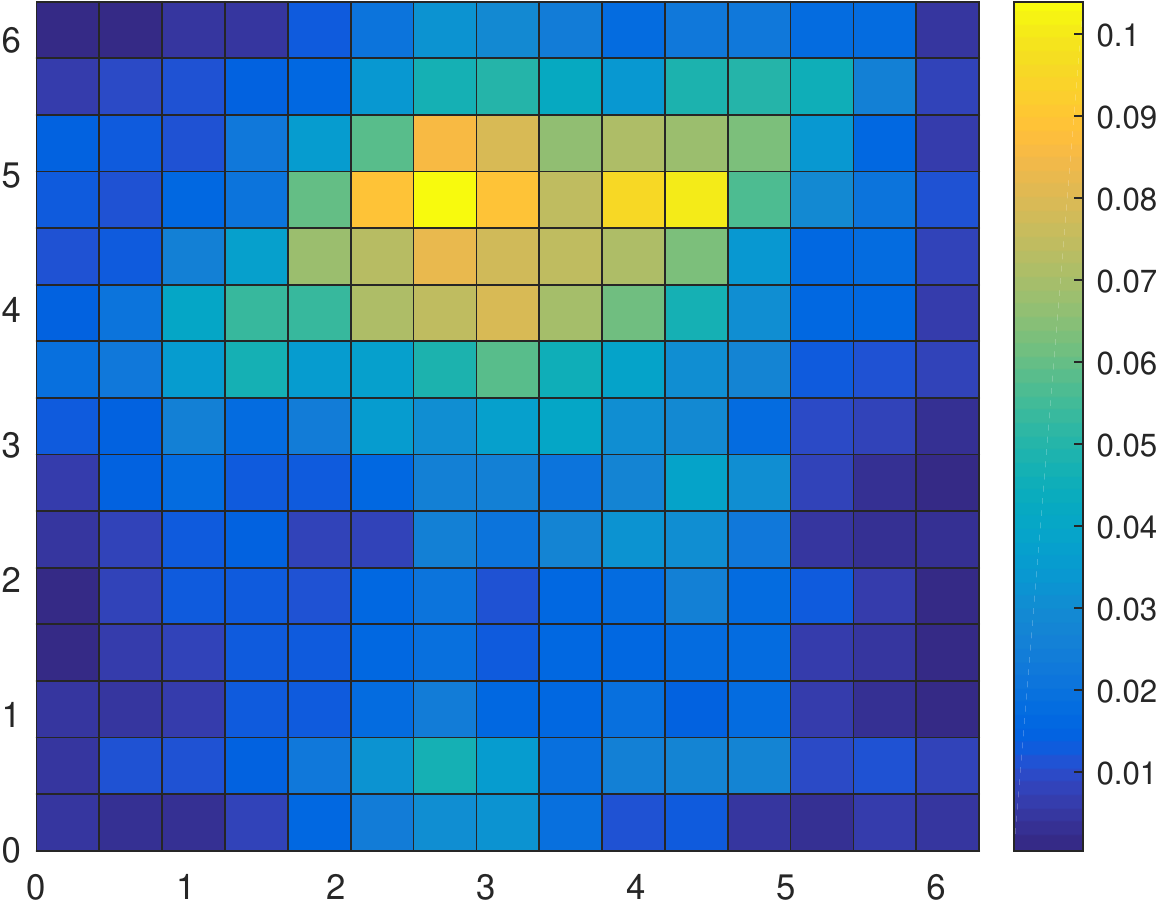}}
		\subfigure[$\kappa=2^{-3}$]{\includegraphics[width=0.24\linewidth]{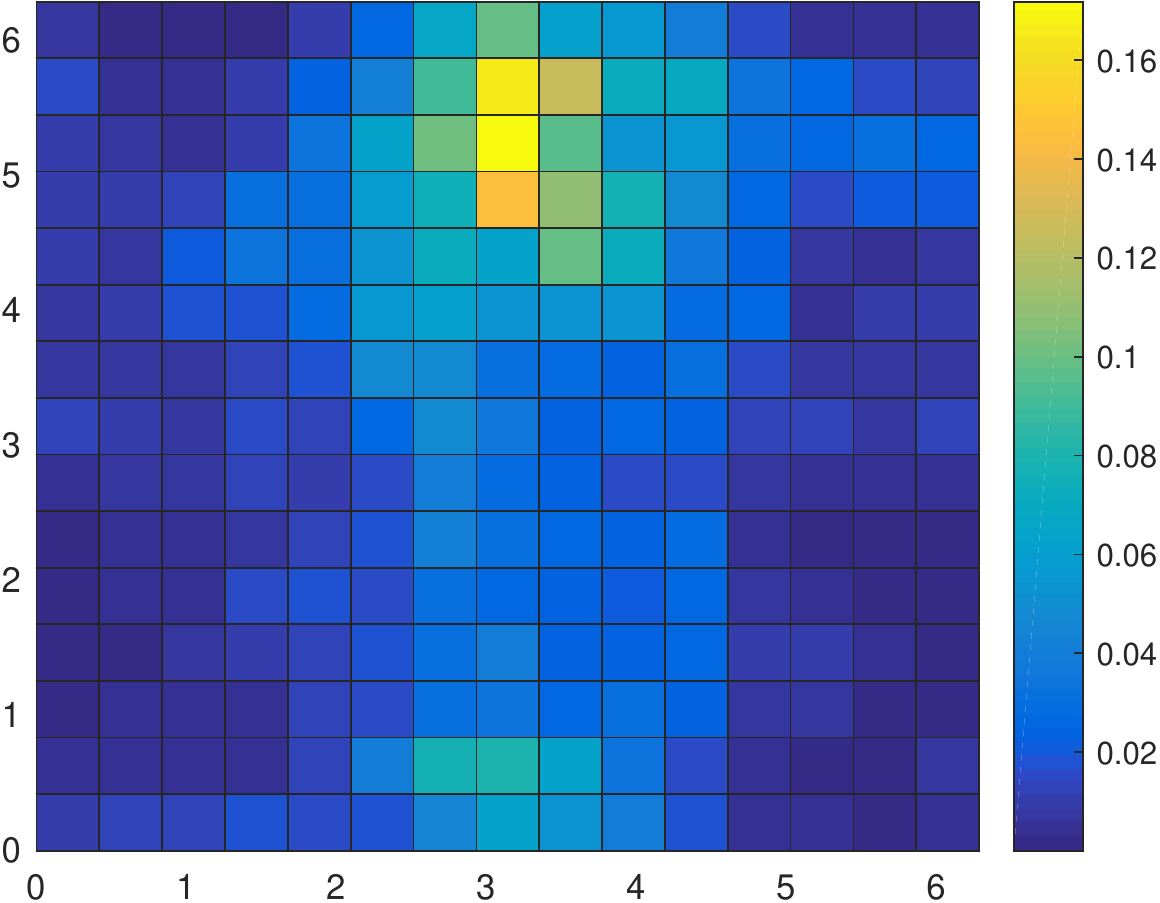}}
			\subfigure[$\kappa=2^{-3.5}$]{\includegraphics[width=0.24\linewidth]{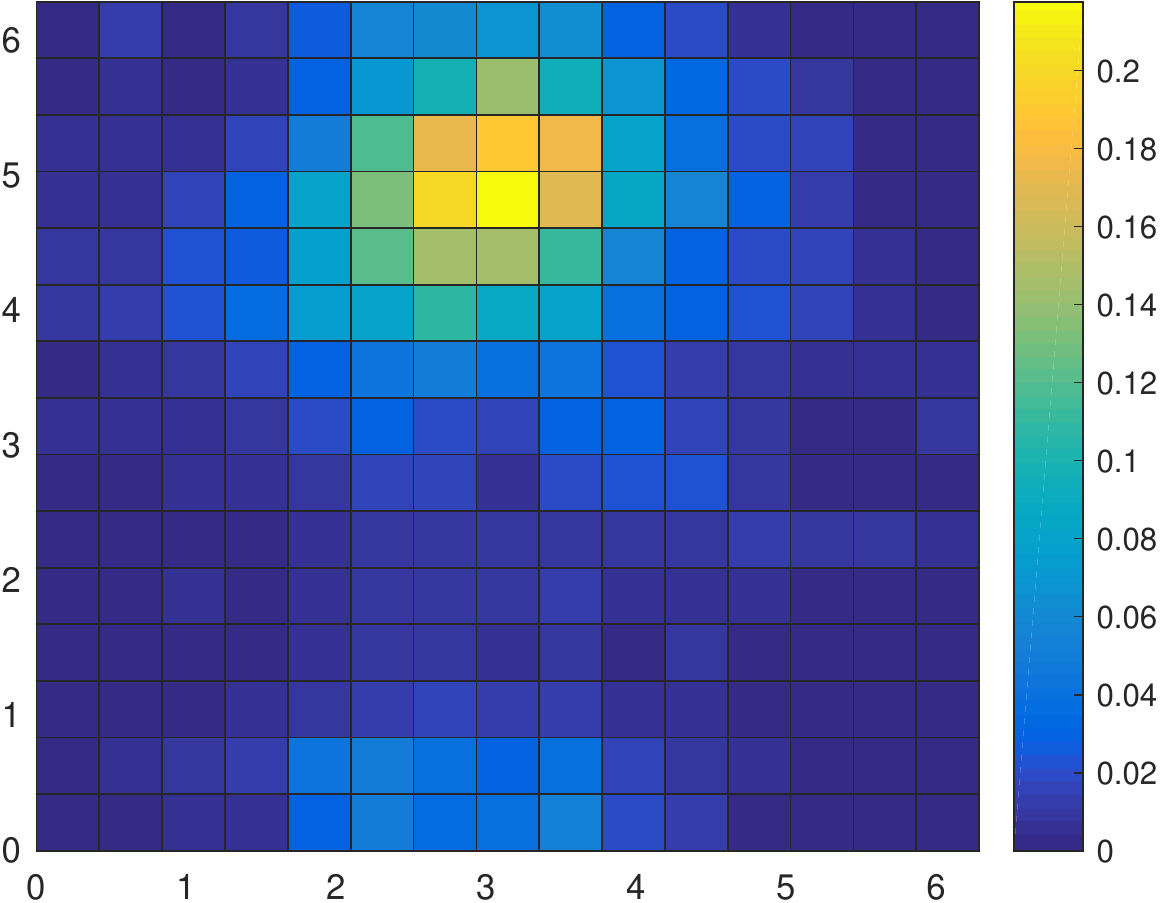}}
	\\
		\subfigure[$\kappa=2^{-2}$]{\includegraphics[width=0.24\linewidth]{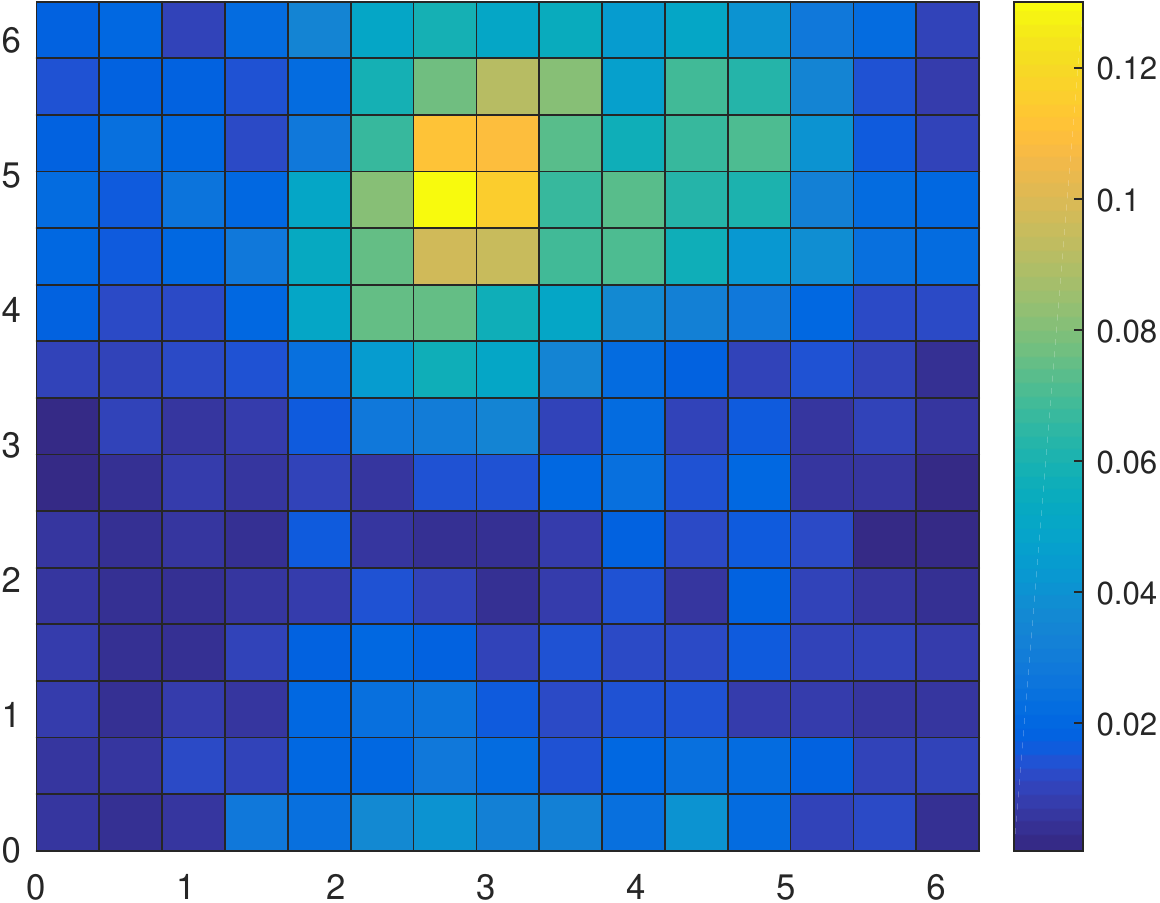}}
	\subfigure[$\kappa=2^{-2.5}$]{\includegraphics[width=0.24\linewidth]{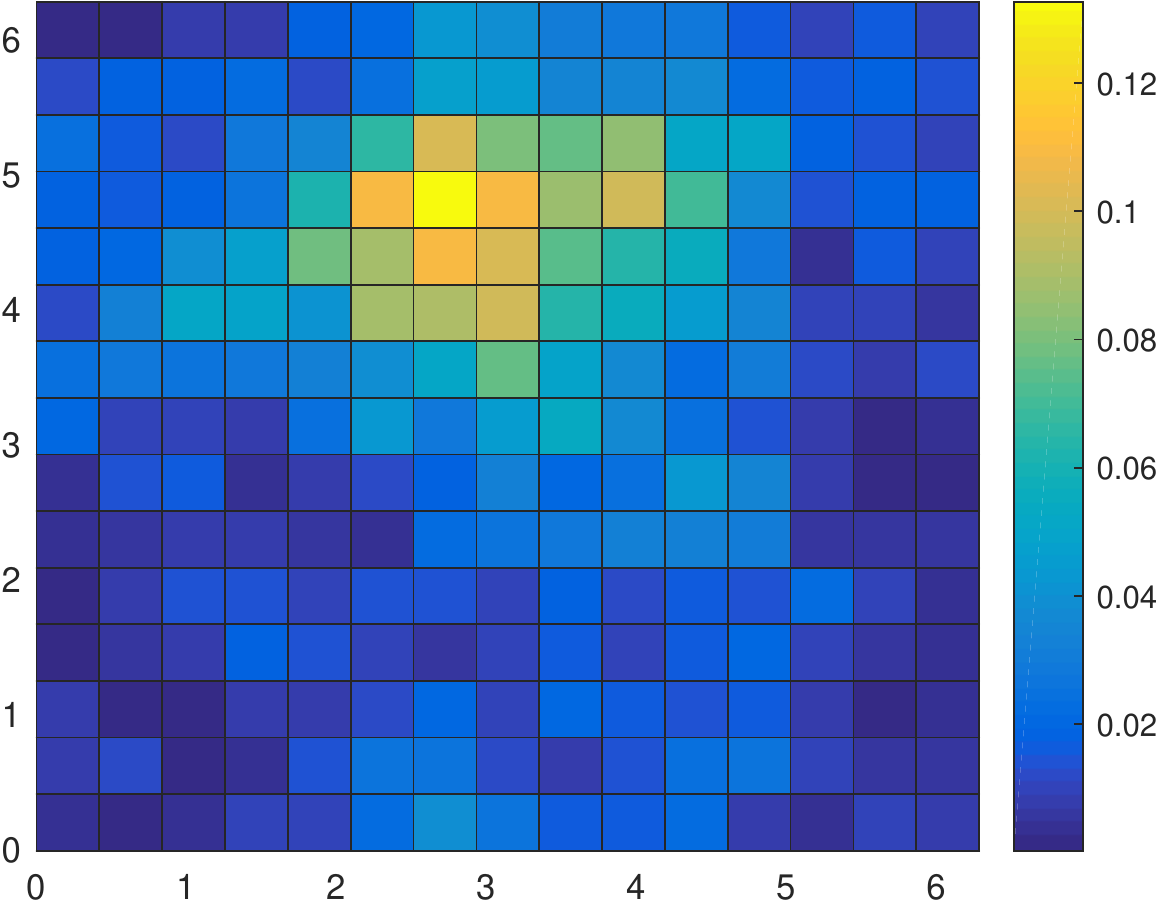}}
	\subfigure[$\kappa=2^{-3}$]{\includegraphics[width=0.24\linewidth]{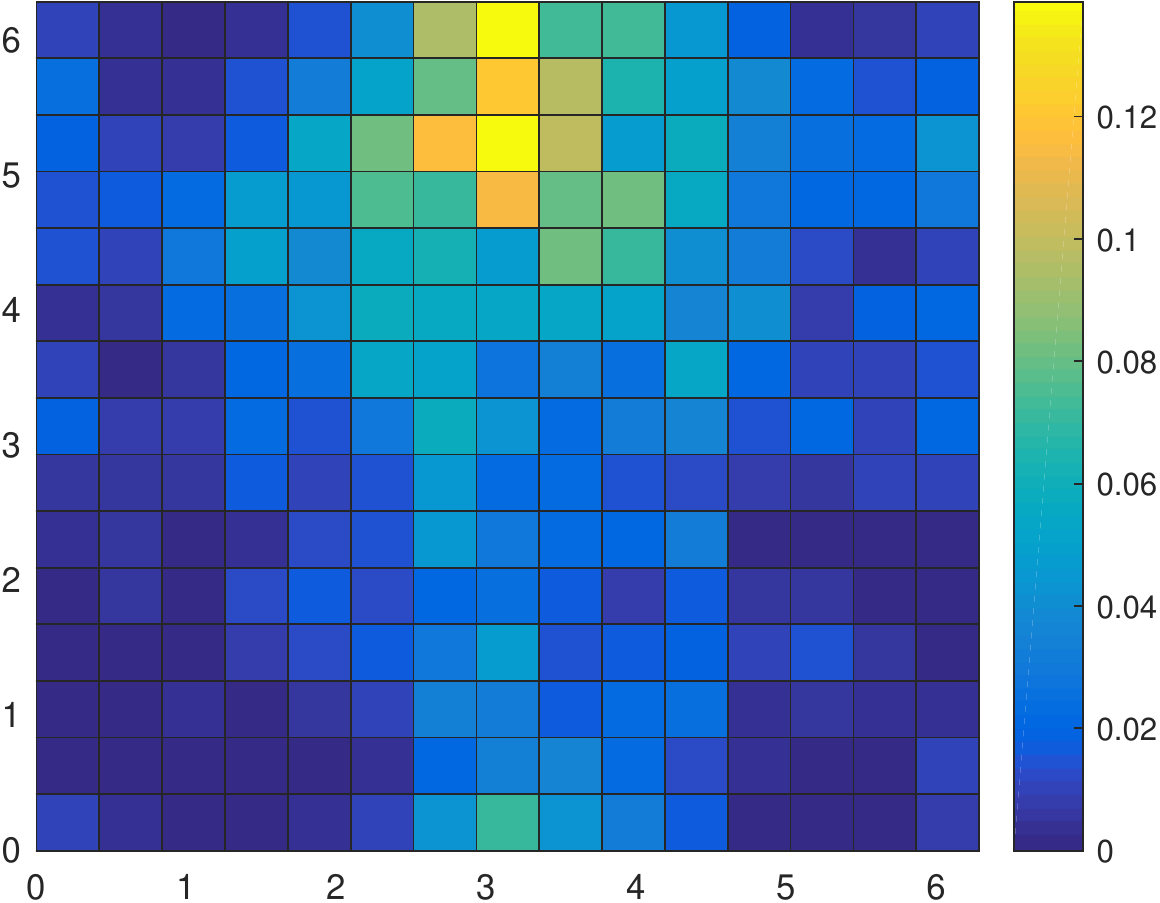}}
	\subfigure[$\kappa=2^{-3.5}$]{\includegraphics[width=0.24\linewidth]{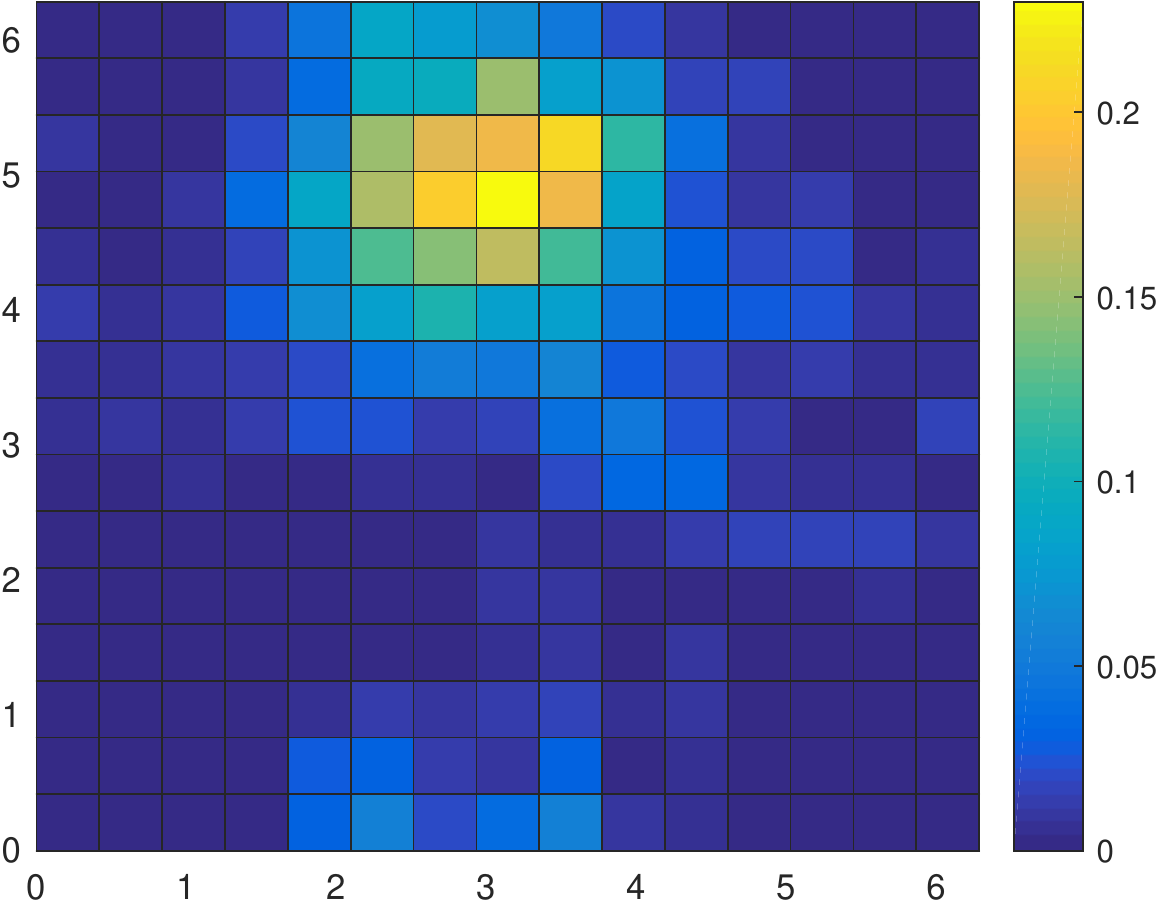}}
	\caption{Generated invariant measure (upper row) and the corresponding training data (reference, lower row), viewed as projections to 2nd and 3rd dimensions in a 3D KPP front speed computation.}
	\label{fig:eg3d_kappa}
\end{figure}

\begin{figure}[htbp]
	\centering
	\subfigure[Ground truth invariant measure]{\includegraphics[width=0.44\linewidth]{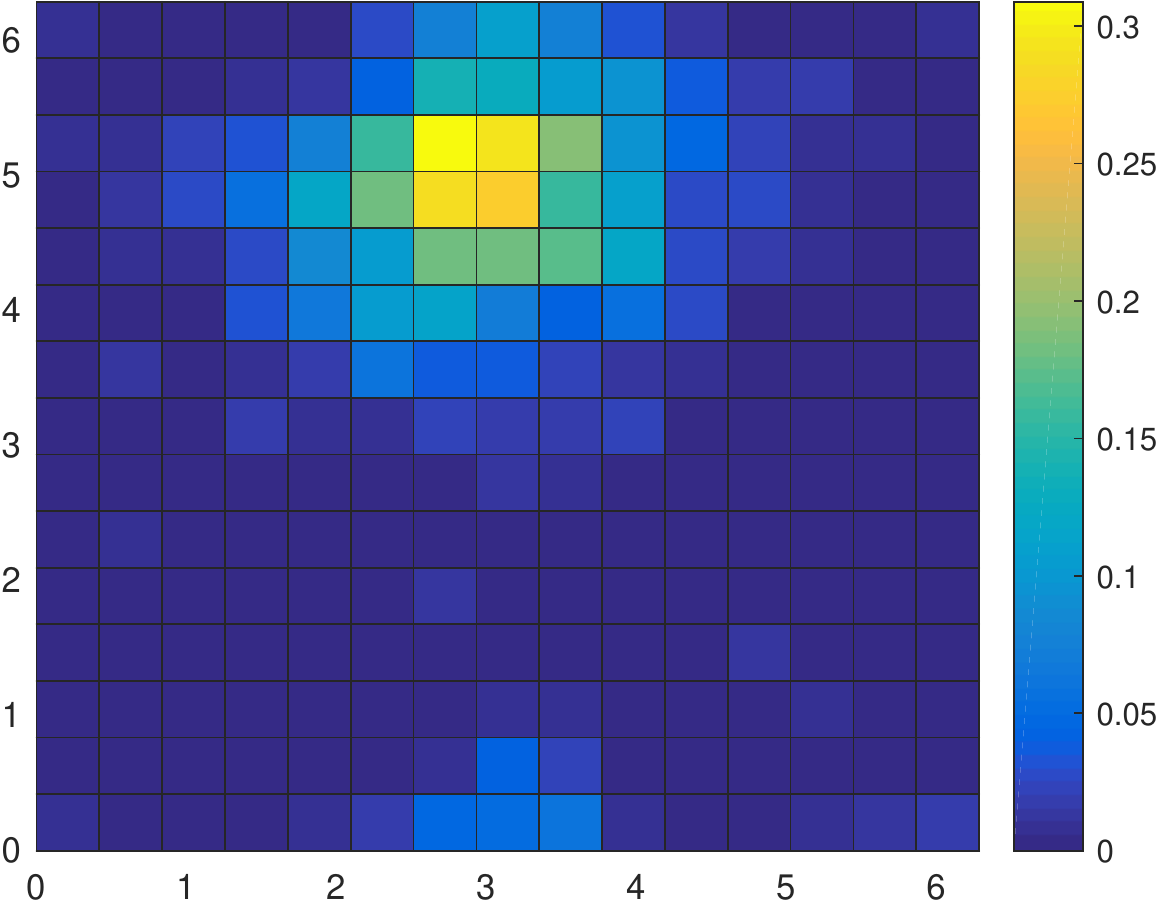}}
	\subfigure[Predicted invariant measure]{\includegraphics[width=0.44\linewidth]{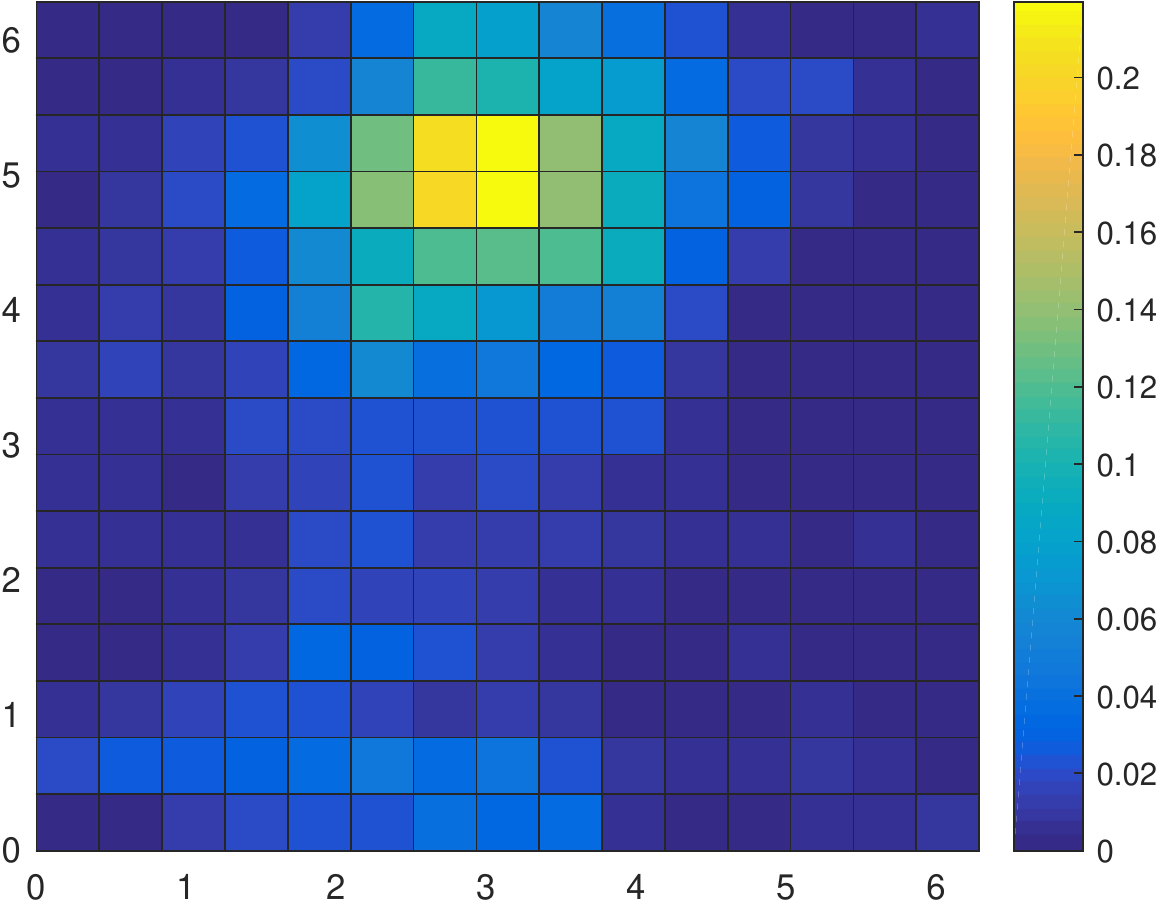}}\\
	\subfigure[$W^2$ training error vs. gradient descent steps: spikes with height $\approx 0.2$ occur due to transition matrix $\gamma$ re-optimized in response to mini-batching of input training data.]{\includegraphics[width=0.44\linewidth]{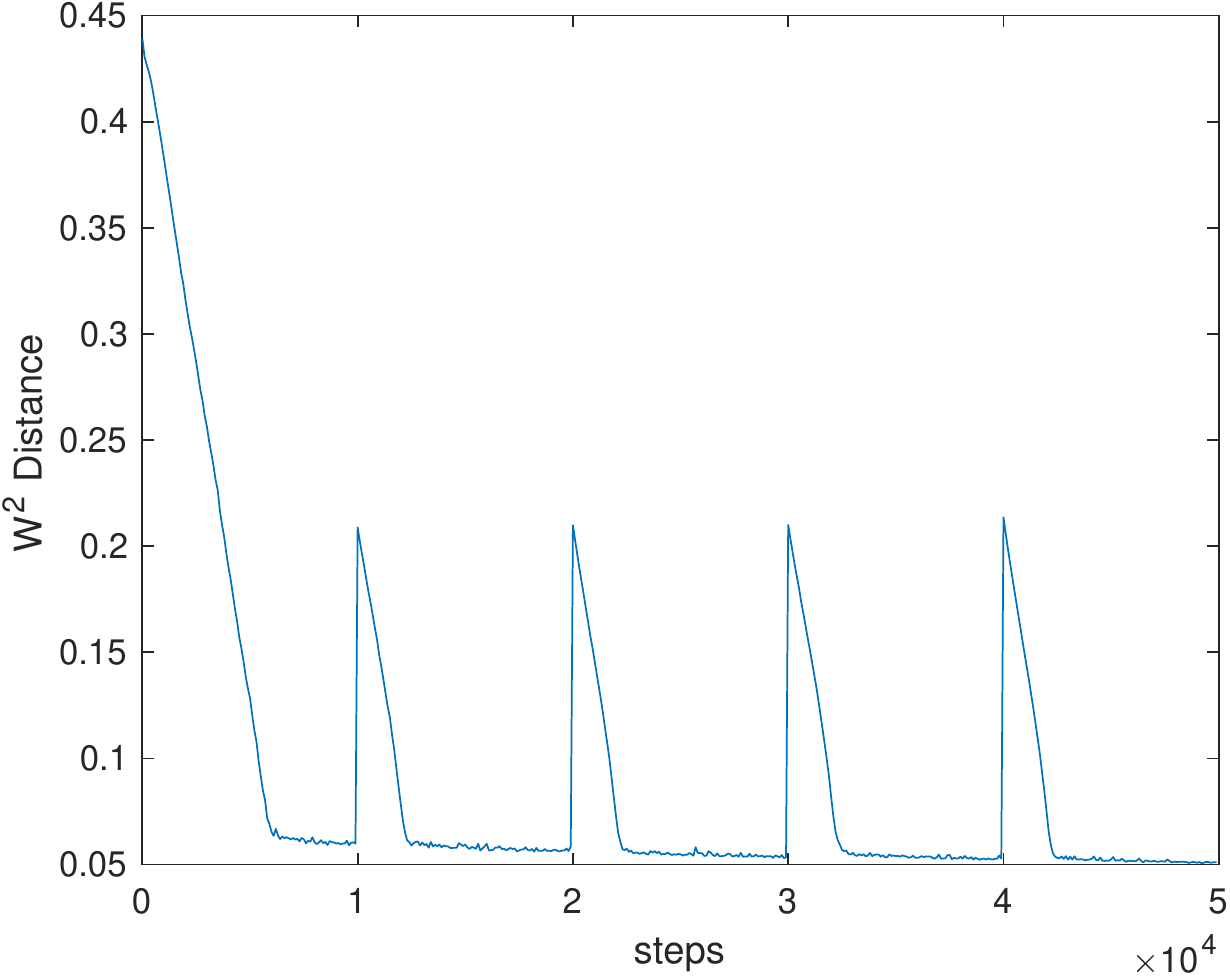}}
	\subfigure[Acceleration: convergence to $\lambda_{\Delta t}$ value computed by 
	Alg.\ref{gIPM1} with warm/cold start by DeepParticle prediction (red)/uniform distribution (blue).]{\includegraphics[width=0.44\linewidth]{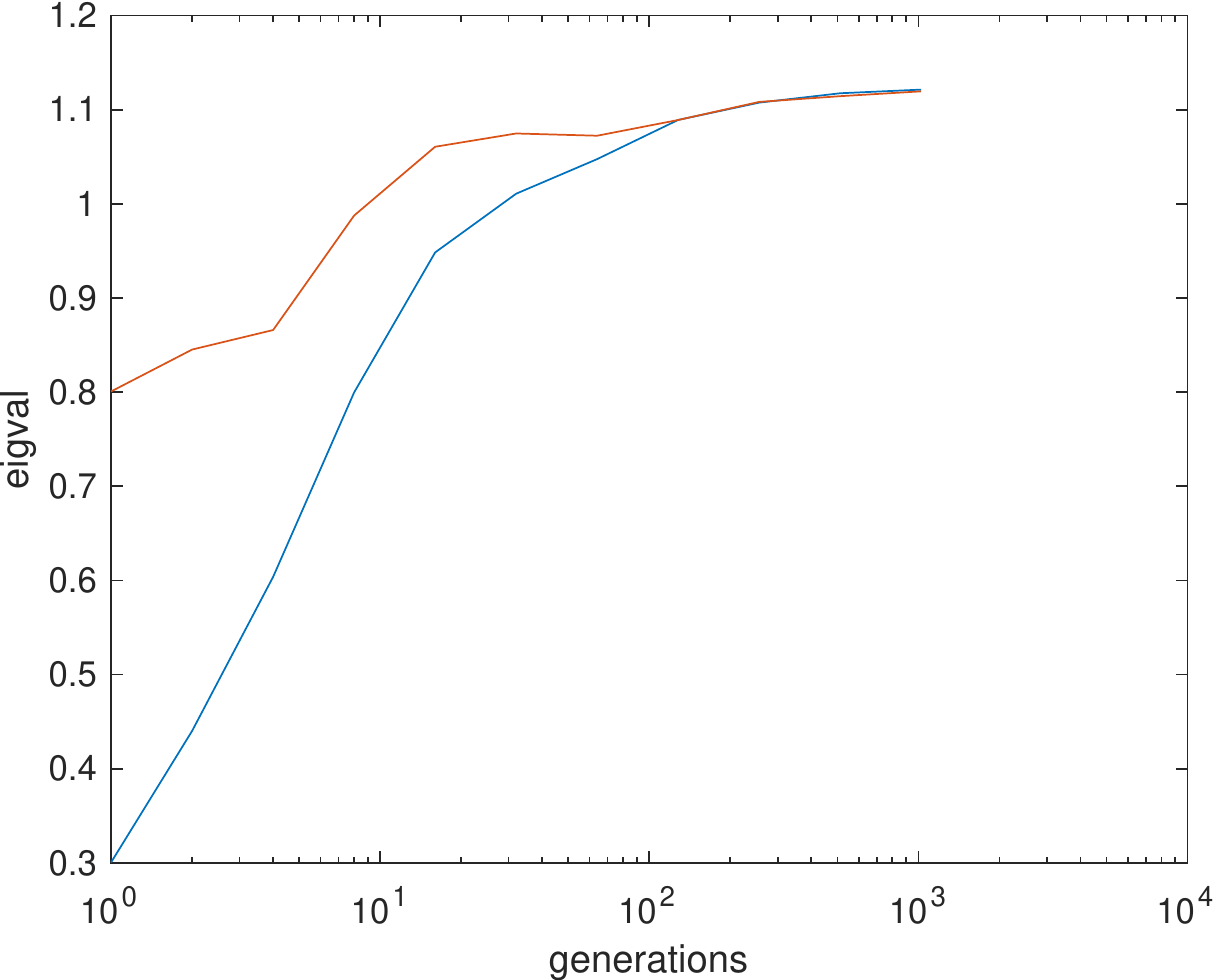}}
	\caption{DeepParticle prediction (viewed in 2nd/3rd dim, top) at test value $\kappa=2^{-4}$ in a 3D computation. 
	The $W^2$ distance minimization in c) shows fast (steps before 1.e4) and slow phases \cite{ZYX_21}.
	}
		\label{fig:eg3d_prediction}
\end{figure}

Fig. \ref{fig:eg3d_kappa} shows the performance of the network interpolating training data from $\kappa=2^{-2}$ to $\kappa=2^{-3.5}$.  Fig. \ref{fig:eg3d_prediction} displays network  prediction at $\kappa=2^{-4}$. All the histograms here are 2D projections of the 3D histogram to the second and third dimensions. Results of the projection to other choices of two dimensions are similar and are not shown here for brevity.

\section{Conclusions}
\noindent
We developed a DeepParticle method to generate invariant measures of stochastic dynamical (interacting particle) systems by a physically parameterized DNN that minimizes the Wasserstein distance between the  source and target distributions. Our method is very general in the sense that we do not require distributions to be in closed-form and the generation map to be invertible. Thus, our method is fully data-driven and applicable in the fast computation of invariant measures of other interacting particle systems. During the training stage, we update network parameters based on a discretized Wasserstein distance defined on finite samples. We proposed an iterative divide-and-conquer algorithm that allows us to significantly reduce the computational cost in finding the optimal transition matrix in the Wasserstein distance. We carried out numerical experiments to demonstrate the performance of the proposed method. Numerical results show that our method is very efficient and helps accelerate the computation of invariant measures of interacting particle  systems for KPP front speeds. In the future, we plan to apply the DeepParticle method to learn and generate invariant measures of other stochastic particle systems.
 
\section*{CRediT authorship contribution statement}
\noindent
\textbf{Zhongjian Wang}: Conceptualization, Programming, Methodology, Writing-Original draft preparation. \textbf{Jack Xin}: Conceptualization, Methodology, Writing-Reviewing and Editing. \textbf{Zhiwen Zhang}: Conceptualization,  Writing-Reviewing and Editing.

\section*{Declaration of competing interest}
\noindent
The authors declare that they have no known competing financial interests or personal relationships that could have appeared to influence the work reported in this paper.

\section*{Acknowledgements}
\noindent
The research of JX is partially supported by NSF grants DMS-1924548 and DMS-1952644. The research of ZZ is supported by Hong Kong RGC grant projects 17300318 and 17307921, National Natural Science Foundation of China  No. 12171406, Seed Funding Programme for Basic Research (HKU), and Basic Research Programme (JCYJ20180307151603959) of The Science, Technology and Innovation Commission of Shenzhen Municipality.

\bibliographystyle{plain}
\bibliography{ZWpaper_DeepLearningPDEDiscCoef} 
\end{document}